\documentclass[10pt,journal,compsoc]{IEEEtran}
\ifCLASSOPTIONcompsoc
  \usepackage[nocompress]{cite}
\else
  \usepackage{cite}
\fi
\ifCLASSINFOpdf
\else
\fi
\usepackage{graphicx}
\usepackage{amsmath}
\usepackage{amssymb}
\usepackage{booktabs}
\usepackage{multirow}
\usepackage{algorithm}
\usepackage{algorithmic}
\usepackage{color}
\usepackage{amsthm}
\usepackage{amsfonts}
\usepackage{bbold}
\usepackage{hyperref}
\usepackage{subcaption}
\usepackage{amsthm}
\usepackage{chngcntr}
\usepackage{makecell}
\newcommand{\revised}[1]{{\color{black}#1}}

\newtheorem{proposition}{Proposition}
\newtheorem{definition}{Definition}

\newtheorem{theorem}{Theorem}
\newtheorem{corollary}{Corollary}

\usepackage[capitalize]{cleveref}
\crefname{section}{Sec.}{Secs.}
\Crefname{section}{Section}{Sections}
\Crefname{table}{Table}{Tables}
\crefname{table}{Tab.}{Tabs.}

\begin{document}
%
\title{Label Hierarchy Transition: Delving into \\ Class Hierarchies to Enhance Deep Classifiers}
%
%
%
%

\author{Renzhen~Wang,
        De~Cai,
        Kaiwen~Xiao,
        Xixi~Jia,
        Xiao~Han,~{Senior Member,~IEEE}
        Deyu~Meng,~{Member,~IEEE}
\IEEEcompsocitemizethanks{
\IEEEcompsocthanksitem Renzhen~Wang is with School of Mathematics and Statistics and Ministry of Education Key Lab of Intelligent Networks and Network Security, Xi'an Jiaotong University, Xi'an, China (e-mail: rzwang@xjtu.edu.cn).

\IEEEcompsocthanksitem De~Cai was with Tencent AI Lab, Shenzhen, China. He is now with ByteDance, Shenzhen, China (e-mail: caide199212@gmail.com).

\IEEEcompsocthanksitem Kaiwen Xiao was with Tencent AI Lab, Shenzhen, China. He is now with SingPath AI Lab, SingPath Medical Technology Pte. Ltd., Singapore (e-mail: kevin.xiao@singpath.com).

\IEEEcompsocthanksitem Xixi~Jia is with School of Mathematics and Statistics, Xidian University, Xi'an, China (e-mail: hsijiaxidian@gmail.com).

\IEEEcompsocthanksitem Xiao~Han was with Tencent AI Lab, Shenzhen, China. He is now with the College of Biomedical Engineering, Sichuan University, Chengdu, China (e-mail: xiao\_han@scu.edu.cn).

\IEEEcompsocthanksitem Deyu Meng (corresponding author) is with the School of Mathematics and Statistics, Ministry of Education Key Lab of Intelligent Networks and Network Security, Xi'an Jiaotong University, Xi’an, China, and also with the Macao Institute of Systems Engineering, Macau University of Science and Technology, Taipa, Macao (e-mail: dymeng@mail.xjtu.edu.cn).
\vspace{1mm}}
}

%
%

\markboth{Journal of \LaTeX\ Class Files,~Vol.~14, No.~8, August~2015}%
{Shell \MakeLowercase{\textit{et al.}}: Bare Demo of IEEEtran.cls for Computer Society Journals}
%



\IEEEtitleabstractindextext{%
\begin{abstract}
Hierarchical classification aims to sort the object into a hierarchical structure of categories. For example, a bird can be categorized according to a three-level hierarchy of order, family, and species. Existing methods commonly address hierarchical classification by decoupling it into a series of multi-class classification tasks. However, such a multi-task learning strategy fails to fully exploit the correlation among various categories across different levels of the hierarchy. In this paper, we propose Label Hierarchy Transition (LHT), a unified probabilistic framework based on deep learning, to address the challenges of hierarchical classification. The LHT framework consists of a transition network and a confusion loss. The transition network focuses on explicitly learning the label hierarchy transition matrices, which has the potential to effectively encode the underlying correlations embedded within class hierarchies. The confusion loss encourages the classification network to learn correlations across different label hierarchies during training. The proposed framework can be readily adapted to any existing deep network with only minor modifications. We experiment with a series of public benchmark datasets for hierarchical classification problems, and the results demonstrate the superiority of our approach beyond current state-of-the-art methods. Furthermore, we extend our proposed LHT framework to the skin lesion diagnosis task and validate its great potential in computer-aided diagnosis. The code of our method is available at \href{https://github.com/renzhenwang/label-hierarchy-transition}{https://github.com/renzhenwang/label-hierarchy-transition}.
\end{abstract}

}

\maketitle

\IEEEdisplaynontitleabstractindextext

%
\IEEEpeerreviewmaketitle

\IEEEraisesectionheading{\section{Introduction}\label{sec:introduction}}

%
%
%
%



\begin{figure*}[t]
    \centering
    \includegraphics[width=1\linewidth]{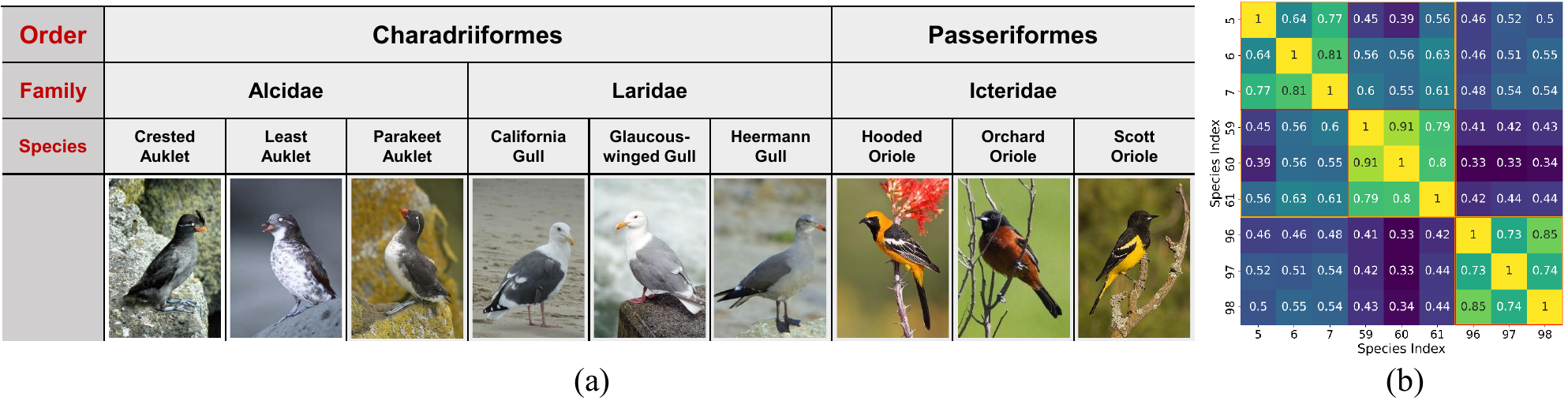}
    \vspace{-4mm}
    \caption{Hierarchical structure in CUB-200-2011 \cite{wah2011caltech}. (a) Example taxonomy (\textit{Species–Family–Order}) with representative bird images, where visual similarity is preserved within hierarchies. (b) Pairwise similarity matrix computed from 312 attributes, showing high correlations within hierarchies and obvious distinctions across coarser-level ones.}
    \label{fig:hierarchy}\vspace{-3mm}
\end{figure*}

\IEEEPARstart{I}{mage} classification is a classic computer vision task that aims to predict a category label for each image. Recent years have witnessed the tremendous success of image classification due to the rapid development of deep learning \cite{lecun2015deep}, especially for deep classification networks \cite{krizhevsky2012imagenet,simonyan2014very,he2016deep}.  Conventionally, a classification network is trained with cross-entropy loss on one-hot labels, such that it predicts almost orthogonal category labels for images in different classes. Such a scheme focuses on the differences of categories, yet it may encounter challenges in effectively capturing inter-class similarity \cite{szegedy2016rethinking,muller2019does}. Hierarchical classification, contrast to classic image classification, attempts to classify the object into a hierarchy of categories, with each hierarchy represents one specific level of concept abstraction. For example, birds can be categorized into a three-level hierarchy of order, family and species according to their taxonomic characters, as shown in Fig. \ref{fig:hierarchy}. In hierarchical classification learning, the labels can be organized according to the domain knowledge (e.g, WordNet database \cite{miller1990introduction}) or automatically generated by algorithms \cite{bannour2012hierarchical,li2010building,marszalek2008constructing,sivic2008unsupervised,wang2019deep}. 

Compared to traditional multi-class classification, hierarchical classification pays more attention to exploit rich semantic correlation among categories \cite{deng2010does}, in order to improve the performance of traditional classification tasks \cite{babbar2013flat,babbar2016learning} or mitigate the severity of prediction mistakes \cite{wu2016learning,bertinetto2020making}, i.e., making the incorrectly classified samples fall within semantically related categories. Take the example of making diagnosis for one malignant tumor,  mistaking it as another malignant one in the same subtype should be much less serious than mistaking it as one benign tumor, as such a mistake would have crucial implications in terms of the follow-up diagnosis and treatment.

Hierarchical classification has been proved effective in many visual scenarios, which can be traced back to traditional linear classifiers \cite{zweig2007exploiting,fergus2010semantic,zhao2011large,liu2013probabilistic}. Recently, hierarchical classification has attracted much attention in deep learning \cite{deng2014large}. Most of the existing works \cite{yan2015hd,wu2016learning,bilal2017convolutional,chen2018fine,chang2021your} commonly fall in a multi-task learning manner, which employs a hierarchy-agnostic feature extractor to learn universal representations across hierarchies and parameterizes a set of hierarchy-specific heads to encode private representations for each hierarchy. For example, a deep network was proposed in \cite{yan2015hd} to classify coarse- and fine-level categories through two different branches. In \cite{wu2016learning}, a granular network was proposed for food image classification, where a single network backbone was shared by multiple fully-connected layers with each one responsible for the label prediction within one hierarchy. Chen et al. \cite{chen2018fine} likewise employed a multi-head architecture to output hierarchy-specific prediction, but instead introduced an attention mechanism to guide label prediction of fine-level hierarchies through that of the coarse-level hierarchies. Furthermore, Alsallakh et al. \cite{bilal2017convolutional} used a standard deep architecture to fit the finest-level labels, and added branches at the intermediate layers to fit the coarser-level labels. Recently, Chang et al. \cite{chang2021your} pointed out that fine-level features benefit coarse-grained classifier learning, yet coarse-level label prediction is harmful to fine-level feature learning. They in turn proposed to disentangle coarse- and fine-level features by leveraging hierarchy-specific classification heads, and further force finer-level features (with stopped gradient) to participate in coarser-level label predictions. Such a simple method has achieved significant improvement in fine-level image recognition. 

Despite significant progress, these recent methods predict category labels for each hierarchy using independent classification heads. This implies that all class hierarchies are statistically independent (to be elaborated in \cref{sec:method}), such that category correlations across hierarchies are encoded solely by a shared feature extractor. Such a multi-task learning strategy is often sub-optimal, as it may lead to the common issue of under-transfer as knowledge can not be sufficiently transferred across different classification heads \cite{long2017learning}. Moreover, these recent methods primarily focus on a restrictive setting where labels for all hierarchies are provided during the training phase. In contrast, the more challenging yet realistic setting, where samples are labeled at only partial hierarchies, remains under-explored \cite{chen2022label}.

To address the aforementioned issues, this paper aims to develop a deep hierarchical classification method capable of \textit{explicitly} capturing category correlations embedded in class hierarchies. Unlike prior work that \textit{implicitly} encodes category correlations through a universal feature extractor, our approach explicitly connects category predictions between adjacent hierarchies. We achieve this by using transition probabilities of category labels from one hierarchy to the next, based on the input image or features. Concretely, we propose Label Hierarchy Transition (LHT), a unified probabilistic approach, to recursively predict the labels from fine- to coarse-level class hierarchies. The LHT model contains two components: (1) an arbitrary existing classification network, which is used to predict the labels for the finest-level hierarchy; (2) a transition network, which generates a label hierarchy transition matrix for each hierarchy (except for the finest-level one). Each element of the matrix denotes the transition probability of one class at the corresponding hierarchy given the other class of the adjacent finer-level hierarchy. For each coarser-level hierarchy, the category prediction can be formulated as the product of its corresponding transition matrix and the prediction score of its adjacent finer-level hierarchy. The two components are jointly trained in an end-to-end manner, and the learned transition matrices for coarser-level hierarchies theoretically provide supervision for the finer-level ones. This enables our proposed LHT framework significantly improve the performance of hierarchical classification, especially in scenarios where a large number of samples have labels only at partial hierarchies.

Moreover, the hierarchical cross-entropy loss used in prior works based on multi-task learning tends to produce orthogonal predictions at each hierarchy and is not conducive to the learning of correlations among different categories. To address this challenge, we further propose a confusion loss to explicitly regularize the label hierarchy transition matrices, which facilitates the network to learn category correlations from all class hierarchies during the training phase.

 In summary, the main contributions are four-fold:
\begin{itemize}
    \item We propose a unified deep classification framework to address the hierarchical classification problem. Contrast to predicting the labels for each hierarchy, we propose to learn label hierarchy transition matrices, which encode the likelihood of similarity between classes across different label hierarchies. To the best of our knowledge, we are the first to utilize deep networks to explicitly learn transition probabilities across class hierarchies.
    \item We provide grounded theoretical understanding that explains how our method potentially works in real-world scenarios. Specifically, we prove that the proposed method facilitates bi-directional information flow between fine- and coarse-level hierarchies, enabling both granular detail enhancement and broader contextual integration. Furthermore, the learned transition matrices at coarse-level hierarchies serve as supervisory signals for fine-level ones, offering a practical advantage in missing-label scenarios.
    \item We propose a confusion loss to directly regularize the negative entropy of the column space of label hierarchy transition matrices, which further encourages the network to learn correlations among class hierarchies, and alleviates the issue of over-confident predictions arising from cross-entropy loss.
    \item Our LHT method can be readily applied to any existing classification network, and the resulting model can be trained in an end-to-end manner. Extensive experiments on a series of hierarchical classification tasks demonstrate the superiority of our approach beyond the prior state-of-the-arts.
\end{itemize}

The rest of this paper is organized as follows. Section \ref{sec:related} revisits related works. Section \ref{sec:method} describes the formulation of the proposed LHT framework as well as its training algorithm. Section \ref{sec:experiment} presents experimental results and analysis in detail. Section \ref{sec:extension} provides a realistic application of our method in computer-aided diagnosis. Section \ref{sec:conclusion} summarizes the final conclusion.

\section{Related Work}
\label{sec:related}
Hierarchical classification \cite{gordon1987review,tousch2012semantic,silla2011survey}, exploiting a hierarchical structure of labels for classification task, has been extensively explored in natural language processing \cite{lewis2004rcv1,mayne2009hierarchically,rousu2006kernel}, computer vision and computer-aided diagnosis \cite{esteva2017dermatologist,yu2022skin,yang2020hierarchical}.
In the existing literature, hierarchical classification is mainly related to three aspects of research: hierarchical architecture, hierarchical loss function, and label embedding based methods. Next, we discuss related studies with respect to these three aspects.

\subsection{Hierarchical architecture}
A common way of designing a hierarchical architecture is to alleviate the computational burden of traditional linear classifiers when the number of recognition categories become huge \cite{griffin2008learning,bengio2010label,gao2011discriminative,deng2011fast,liu2013probabilistic}. The core idea was to create a tree-like classifier whose leaf nodes were used for the prediction of the given categories and internal nodes for the cluster of categories.

Most recently, deep learning based methods are increasingly used to address hierarchical classification \cite{yan2015hd,wu2016learning,cerri2016reduction,bilal2017convolutional,chen2018fine,chang2021your,chen2022label}. These methods commonly fall into a multi-task learning manner, and adopt shared-specific feature representation learning to fit labels across different levels of hierarchy. Such a flexible architecture is advantageous in learning low-level features on one hand, while on the other hand, it can easily incorporates canonical inclusion relations between the parent node (super-class) and its child nodes (sub-classes) \cite{cerri2016reduction,chang2021your} or exclusion relations \cite{chen2022label} between the sibling nodes at the same hierarchy into the network design. Contrary to these works, our proposed LHT framework learns the conditional probabilities between two arbitrary nodes within adjacent hierarchies to encode the affinity relations.

\begin{figure*}[t]
    \centering
    \includegraphics[width=1.0\linewidth]{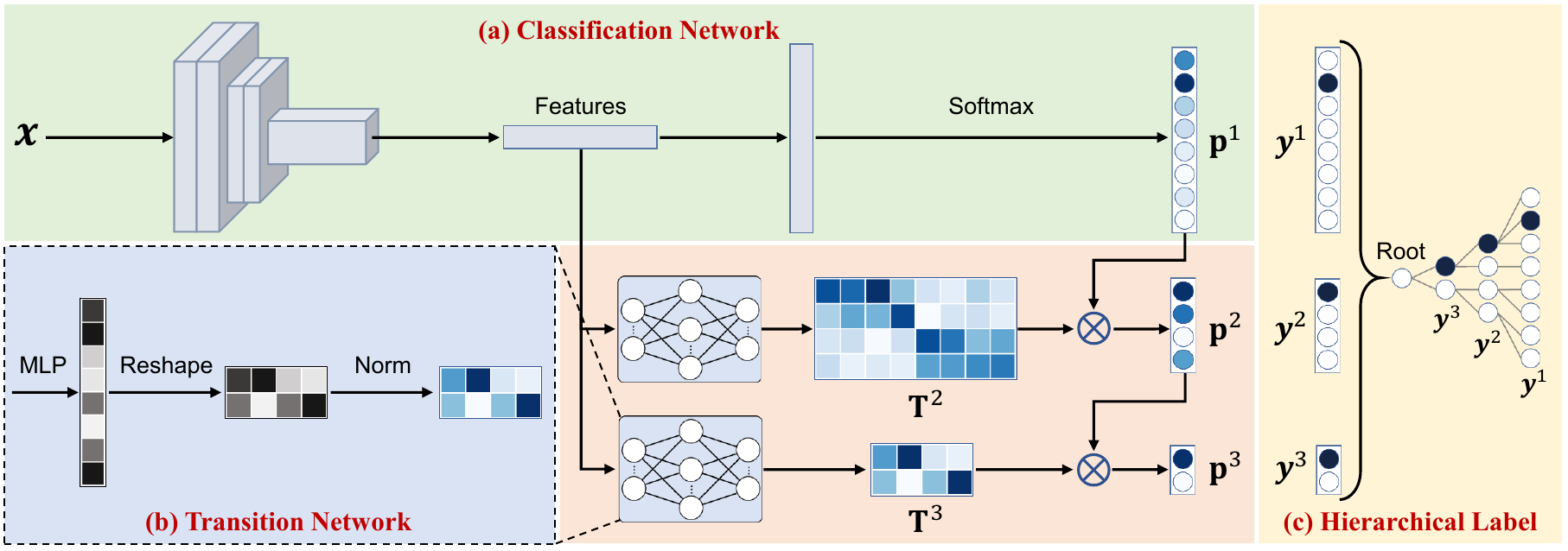}
    \caption{Schematic illustration of the proposed LHT model. We take a three-level hierarchical classification as an example, where the number of classes within each hierarchy is 8, 4 and 2, respectively. The model consists of two key components: (\romannumeral1) Classification Network, which can be any existing classification network for predicting the labels of the finest-level hierarchy; (\romannumeral2) Transition Network, which leverages a light-weight network (e.g., a multi-layer perceptron (MLP) with one single hidden layer) to map the feature embedding of the input sample to the label hierarchy transition matrix for each hierarchy. For each coarse-level hierarchy, its label prediction is obtained by multiplying the corresponding transition matrix by the label prediction of its adjacent fine-level hierarchy.}
    \label{fig:framework}
\end{figure*}

\subsection{Hierarchical loss function}
A range of literature on hierarchical classification resorts to incorporating the class hierarchies in the loss function. In \cite{verma2012learning}, the metrics associated with each node of the taxonomy tree were learned based on a probabilistic nearest-neighbour classification framework. In \cite{srivastava2013discriminative}, the tree-structure hierarchies were used to impose prior on the parameters of the classification layer for knowledge transfer between classes, and similar strategy was proposed by Zhao et al. \cite{zhao2011large} in the traditional linear classifier. In \cite{deng2014large}, various label relations (e.g., mutual exclusion and subsumption) were encoded as conditional random field to pair-wisely capture the correlation between classes. In \cite{giunchiglia2020coherent}, max constraint loss was proposed to produce predictions coherent with the hierarchy constraint for multi-label hierarchical classification. Recently, Bertinetto et al. \cite{bertinetto2020making} proposed a hierarchical loss that could be viewed as a weighted cross-entropy loss, and the weights were defined according to the depth of the label tree. In parallel to our method, chen et al. \cite{chen2022label} proposed a probabilistic loss to encode the label relation within each hierarchy.  Unlike these methods \cite{hssn,hssn+}, we combine cross-entropy with a novel confusion loss to regularize the learned transition matrices, thereby fully exploiting class correlations across hierarchies. Besides, in hierarchical segmentation \cite{hssn,hssn+,weber2024flattening}, prior works such as HSSN \cite{hssn,hssn+} incorporates class hierarchies by directly constraining the output space through a tree-min loss to produce hierarchy-coherent predictions. In contrast, our LHT takes a probabilistic approach by learning transition matrices to model inter-class similarities, which not only enforces hierarchy coherence, but also leverages cross-level correlations more flexibly, especially under missing-label scenarios.

\subsection{Label embedding}
Label embedding aims to encode the correlation information among classes into the traditional one-hot labels, which has been widely explored in multi-class classification \cite{szegedy2016rethinking,chorowski2016towards,muller2019does,guo2021label}. For example, Label Smoothing presented in \cite{szegedy2016rethinking,muller2019does} softens the one-hot label by assigning a slightly lower probability to the target class and distributes the remaining probability evenly across the non-target classes. In hierarchical classification, label embedding techniques typically map one-hot labels to vectors, where the relative positions of these vectors encode the structural information embedded in the class hierarchy. Specifically, Bertinetto et al. \cite{bertinetto2020making} proposed Soft Labels, an embedding-based method that softens one-hot labels using a metric derived from the lowest common ancestor (LCA) height within a given label tree. Dhall et al. \cite{dhall2020hierarchical} employed entailment cones to learn order-preserving label embeddings and subsequently integrated them with image embeddings for hierarchical classification by using the transitivity of the embeddings across levels in the hierarchy. Although label embedding is a potential strategy to capture the correlation across class hierarchies, in this paper, we attempt to develop a unified deep classification framework for explicitly encoding correlation of hierarchical classification.

\section{The Proposed Method}
\label{sec:method}
In this section, we first introduce the problem formulation of hierarchical classification and the preliminaries of our work. We then provide an overview of our proposed LHT framework that explicitly encodes category correlations across distinct hierarchies. Next, we present the architecture of the proposed transition network. Finally, we introduce the proposed confusion loss to encourage correlation learning across hierarchies.

\subsection{Problem Formulation and Preliminaries}
\label{subsec:preliminary}
In hierarchical classification tasks, a given image $\mathbf x$ is typically annotated with a sequence of $K$ labels $(y^1, y^2, \ldots, y^K)$, where $K$ represents the number of label hierarchies. Each $y^k \in \{1, 2, \ldots, C_k\}$ is the label within the $k$-th hierarchy, with $C_k$ denoting the number of categories of this hierarchy. We denote $\mathbf y^k$ is the corresponding one-hot vector of $y^k$. Without loss of generality, we define $k=1$ as the finest-level hierarchy, and assume the granularity becomes coarser as $k$ increases. Accordingly, the number of classes decreases with larger $k$, i.e., $C_k > C_{k+1}$. The goal of hierarchical classification is to learn a classifier to predict hierarchical labels for each unknown $\mathbf x$.

The multi-task approach to addressing hierarchical classification commonly involves two steps: \romannumeral1) it first employs a hierarchy-agnostic feature extractor $f_\omega$ parameterized by $\omega$ to learn a universal representation $\hat{\mathbf x}=f_\omega(\mathbf x)$ across hierarchies, and \romannumeral2) for the $k$-th hierarchy, it uses a hierarchy-specific classification head $g^k_{\theta^k}$ parameterized by $\theta^k$ to predict the corresponding classification distribution $\mathbf p^k=g^k_{\theta^k}(\hat{\mathbf x})$. Note that here, with some abuse of notation, we define $\mathbf p^k=[p(Y^k=1|\mathbf x),\cdots,p(Y^k=C_k|\mathbf x)]^T$, where $Y^k$ denotes a random variable representing the label of the $k$-th hierarchy. The involved parameters are jointly optimized using a hierarchical cross-entropy loss
\begin{equation}
    \mathcal L_{{\rm CE}}=\sum_{n=1}^N\sum_{k=1}^K H(\mathbf p^k_n, \mathbf y^k_n),
    \label{eq:hierarchical_ce}
\end{equation}
where $\{\mathbf x_n,(\mathbf y_n^1,\cdots,\mathbf y_n^K)\}_{n=1}^N$ represents a batch of $N$ training samples, and $H(\cdot,\cdot)$ is the cross-entropy loss defined as $H(\mathbf p, \mathbf y)=\sum_i \mathbf y[i]\log \mathbf p[i]$, with $\mathbf p[i]$ denoting the $i$-th entry of $\mathbf p$.

\cref{eq:hierarchical_ce} is equivalent to solve the negative log-likelihood for $p(y^1,\cdots,y^K|\mathcal I)=\prod_{k=1}^K p(y^k|\mathcal I)$. The detailed derivation can be found in Appendix A.1. This implies that all class hierarchies are considered statistically independent of each other, such that category correlations across hierarchies are encoded solely by the shared feature extractor $f_\omega(\cdot)$. Such a multi-task learning strategy is often sub-optimal due to the following two reasons: \romannumeral1) It fails to explicitly model the hierarchy/task relationships, which may lead to the common issue of under-transfer as knowledge can hardly be transferred across different classification heads \cite{long2017learning}. \romannumeral2) In missing-label scenarios where most samples are labeled at only partial hierarchies, there is a risk that the classification head associated with the finest-level hierarchy might overfit, as it is trained on fewer samples compared to the coarser-level hierarchies.

In light of the aforementioned reasons, a natural question is how to explicitly model hierarchical relationships within a multi-task deep learning framework to facilitate information transfer across different classification heads, especially in challenging scenarios with missing labels.

\subsection{Label Hierarchy Transition}
\label{subsec:overall}

The goal of this paper is to propose a unified probabilistic framework to explicitly encode category correlations across hierarchies into the existing multi-task learning framework. Unlike previous works based on the assumption of independent distributions across different hierarchies, we mathematically consider that the label $Y^{k}$ of the $k$-th hierarchy is statistically dependent on $Y^{k-1}$ and the sample $\mathbf x$ itself. It is natural to encode such correlations between adjacent hierarchies through the conditional distribution $p(Y^k|Y^{k-1},\mathbf x)$. Subsequently, we can formulate the distribution $p(Y^k|\mathbf x)$ by marginalizing over the variable $Y^{k-1}$:
\begin{equation}
p(Y^k|\mathbf x)\!=\!\sum_{i=1}^{C_{k-1}}p(Y^k|Y^{k-1}=i,\mathbf x)\cdot p(Y^{k-1}=i|\mathbf x),
\label{eq:transition}
\end{equation}
where $p(Y^{k-1}|\mathbf x)$ is the label distribution of the $(k\!-\!1)$-th hierarchy with $k=2,\cdots,K$.

 For simplicity, here we define
\begin{equation}
t^k_{ij}\triangleq p(Y^k=i|Y^{k-1}=j,\mathbf x),
\label{eq:probability}
\end{equation}
and then we have a $C_k\times C_{k\!-\!1}$ matrix
\begin{equation}
  \mathbf T^k=\begin{bmatrix}
    \begin{array}{cccc}
    t^k_{11}   & t^k_{12}   & \cdots & t^k_{1C_{k-1}}  \\
    t^k_{21}   & t^k_{22}   & \cdots & t^k_{2C_{k-1}}  \\
    \vdots     & \vdots     & \ddots & \vdots          \\
    t^k_{C_k1} & t^k_{C_k2} & \cdots & t^k_{C_kC_{k-1}} \\
    \end{array}
    \end{bmatrix}.
\end{equation}
We call this matrix \textit{Label Hierarchy Transition Matrix} (or abbreviated as \textit{Transition Matrix}), since it defines the probabilities of transitions from fine-grained labels to coarse-grained labels and we can get a glimpse of the \emph{transition} status between the adjacent class hierarchies. Note that the elements of $\mathbf T^k$ are positive and each column sums to 1, i.e., $\sum_{i=1}^{C_{k}}t^k_{ij}=1$ for each $j\in\{1,\cdots,C_{k-1}\}$.

\subsubsection{Model Formulation}
In this paper, the core of our \emph{Label Hierarchy Transition} model is to predict the label distribution $p(Y^k|\mathbf x)$ for each hierarchy, by jointly learning the two kinds of conditional probability involved in Eq. \eqref{eq:transition}. Notably, \cref{eq:transition} inherently defines a fine-to-coarse learning paradigm, as illustrated in Fig. \ref{fig:framework}. Within this paradigm, the finest-level label distribution is first fitted, and the coarser-level ones are subsequently learned, conditioned on the input image and its adjacent finer-level label distribution. Specifically:

\begin{itemize}
    \item For the finest-level hierarchy, we use a vanilla \textit{Classification Network} $f_\omega$ with parameters $\omega$ to predict its label distribution $\mathbf p^1$ as follows:
    \begin{equation}
        \mathbf p^1=\sigma(\mathbf z)=\sigma(f_\omega(\mathbf x)),
        \label{eq:fine_level_prediction}
    \end{equation}
    where $\mathbf p^1\!=\![p(Y^1\!=\!1|\mathbf x), \cdots, p(Y^1\!=\!C_1|\mathbf x)]^T$, $\mathbf z$ denotes the logits of $\mathbf x$, and $\sigma$ is the softmax function.
    \item For the $k$-th level hierarchy with $k=2,\cdots,K$, we design a light-weight \textit{Transition Network} $g^k_{\theta^k}$ parameterized by $\theta^k$ to predict its transition matrix $\mathbf T^k$, such that $\mathbf T^k=g^k_{\theta^k}(\mathbf x)$. Thus, the label distribution $\mathbf p^k$ can be formulated as
    \begin{equation}
    {\mathbf p}^k=\mathbf T^k{\mathbf p}^{k-1}.
    \label{eq:coarse_level_prediction}
\end{equation}
The architectural designs of $g^k_{\theta^k}$ will be elaborated in the next subsection.
\end{itemize} 

From \cref{eq:fine_level_prediction,eq:coarse_level_prediction}, we can get the label distributions of all hierarchies. This allows for the joint optimization of the parameters involved in both the Classification Network and the Transition Networks, i.e., $\{\omega, \theta^2,\cdots,\theta^K\}$, in an end-to-end manner. Given a batch of $N$ training samples $\{\mathbf x_n, (\mathbf y_n^1, \mathbf y_n^2,\cdots,\mathbf y_n^K)\}_{n=1}^N$, the hierarchical cross-entropy loss can be formulated as follows:
\begin{equation}
\begin{aligned}
    \mathcal L_{{\rm CE}}&=\sum_{n=1}^N\sum_{k=1}^KH({\mathbf p}_n^k, \mathbf y^k_n) \\
    &=\sum_{n=1}^N\Bigl(H\bigl({\mathbf p}^1_n,\mathbf y^1_n\bigr) \!+\!\sum_{k=2}^K H\bigl(\mathbf T^k_n{\mathbf p}_n^{k-1}, \mathbf y^k_n\bigr)\Bigr).
\end{aligned}
\label{eq:hierarchical_ce_lht}
\end{equation}
In addition to \cref{eq:hierarchical_ce_lht}, we propose a novel confusion loss to enhance the learning of category correlations across all class hierarchies during the training phase. The details will be discussed in \cref{sec:confusion_loss}.

\subsubsection{Theoretical Analysis.} 
Our LHT framework defined by \cref{eq:fine_level_prediction,eq:coarse_level_prediction} 
offers the following theoretical guarantees: \romannumeral1) LHT facilitates the transition of information from the finest-level hierarchy to the coarser-level ones via forward propagation; \romannumeral2) LHT enables the transition of true label information from coarser-level hierarchies to the finer-level ones through
backpropagation; \romannumeral3) The learned transition matrices for coarser-level hierarchies theoretically provide supervised signals for the finer-level hierarchies, thereby benefiting the training process for missing-label scenarios.

For ease of notation, we only consider the single-sample case ($N=1$) for \cref{eq:hierarchical_ce_lht}, but the derivation for the multi-sample case is straightforward. We first introduce the definition of the transition matrix from the $1$-st hierarchy to the $k$-th hierarchy.
\begin{definition}[]
We define the following matrix 
\begin{equation}
{\mathbf T}^{k\leftarrow1}=\begin{bmatrix}
    \begin{array}{cccc}
    t^{k\leftarrow 1}_{11}   & t^{k\leftarrow 1}_{12}   & \cdots & t^{k\leftarrow 1}_{1C_{1}}  \\
    t^{k\leftarrow 1}_{21}   & t^{k\leftarrow 1}_{22}   & \cdots & t^{k\leftarrow1}_{2C_{1}}  \\
    \vdots     & \vdots     & \ddots & \vdots          \\
    t^{k\leftarrow1}_{C_k1} & t^{k\leftarrow1}_{C_k2} & \cdots & t^{k\leftarrow1}_{C_kC_{1}} \\
    \end{array}
    \end{bmatrix}
\end{equation}
as the \textbf {transition matrix from the $1$-st hierarchy to the $k$-th hierarchy}, where
$t^{k\leftarrow1}_{ij}=p(Y^k=i|Y^1=j,\mathbf x)$ is the transition probability from the $j$-th class within the $1$-st hierarchy to the $i$-th class within the $k$-th hierarchy.
\end{definition}

Given the definition of ${\mathbf T}^{k\leftarrow1}$, we show that our proposed LHT facilitates the transition of information from the finest-level hierarchy directly to any coarser-level
ones as in the following \cref{thr:forword_transer}.
\begin{proposition}
\label{thr:forword_transer}
In the LHT framework, the label prediction of the $k$-th ($k>1$) hierarchy can be inferred from the finest-level hierarchy as follows:
\begin{equation}
\mathbf p^k={\mathbf T}^{k\leftarrow1} {\mathbf p}^{1},
\end{equation}
where the transition matrix $\mathbf{T}^{k \leftarrow 1}$ is defined as $\mathbf{T}^{k \leftarrow 1} := \mathbf{T}^k \mathbf{T}^{k-1} \cdots \mathbf{T}^2$, which is implemented by the composite function $g^k_{\theta^k} \circ g^{k-1}_{\theta^{k-1}} \circ \cdots \circ g^2_{\theta^2}$.
\end{proposition}
The proposition can be proved by verifying that each $(i,j)$-th element of ${\mathbf T}^{k\leftarrow1}$ represents the transition probability $p(Y^k=i|Y^1=j,\mathbf x)$. We provide the proof in Appendix A.2. \cref{thr:forword_transer} indicates that the label prediction for the $k$-th hierarchy is conditioned on the information derived from all the preceding $k-1$ hierarchies. This, on the one hand, demonstrates that
our LHT framework facilitates the transition of information from fine-level hierarchies to coarse-level ones through the forward propagation of the framework. On the other hand, since LHT optimizes all parameters end-to-end via \cref{eq:hierarchical_ce_lht}, it implies that LHT also benefits the transition of true label information from coarser-level hierarchies to the finest-level one through gradient backpropagation, as illustrated in the following proposition.

\begin{proposition}[]
\label{thr:backward_transfer}
If the loss function in \cref{eq:hierarchical_ce_lht} are minimized by gradient based algorithms (such as SGD), then the loss term $\mathcal L^k\triangleq H(\mathbf p^k, \mathbf y^k)$ enables the propagation of true label information from coarse-level hierarchies to the finest-level one through gradient backpropagation.
Specifically, the gradient of $L^k$ with respect to the logits $\mathbf{z}$ of the finest-level hierarchy is computed as
\begin{equation}
    \label{eq:backward_transfer}
    \mathcal G^k_j := \frac{\partial\mathcal L^k}{\partial \mathbf z[j]} =\bigl(1-\frac{t^{k\leftarrow{1}}_{y^k j}}{\mathbf p^k[y^k]}\bigr)\mathbf p^1[j],
\end{equation}
where $\mathbf z[j]$ denotes the $j$-th entry of $\mathbf z$ and $k=2,\cdots,K$.
\end{proposition}

The proof of \cref{thr:backward_transfer} is provided in Appendix A.3. \cref{eq:backward_transfer} demonstrates that the coarse-level hierarchies influence the finest-level hierarchy by adjusting its logit $\mathbf z$ using the true label $y_k$ from the $k$-the hierarchy in the end-to-end training phase. Intuitively, if $\mathcal{G}_j^k < 0$, reducing the loss $\mathcal L^k$ by updating parameters along the negative gradient direction ($-\mathcal{G}_j^k$) will effectively increase $\mathbf{z}[j]$; conversely, if \(\mathcal{G}_j^k > 0\) the same process will lead to a decrease in $\mathbf{z}[j]$. In fact, the Transition Networks can rectify inaccurate predictions by adjusting the logits of the finest-level hierarchy during training. This can be formalized in the following corollary.

\begin{corollary}
If the network parameters are updated according to \cref{eq:hierarchical_ce_lht} using SGD-based optimizers, then $\mathcal{G}^k_{j^*} < 0$ holds for $j^*=\arg\max_{j} t^{k\leftarrow{1}}_{y^k j}$, which is the subclass most visually similar to $y^k$ as determined by the Transition Networks.
\end{corollary}
This corollary indicates that reducing $\mathcal L^k$ along the direction ($-\mathcal{G}^k_{j^*}$) can guide the logits $\mathbf z$ of the finest-level hierarchy towards the value that yield higher probability for the subclass most visually similar to $y^k$ during training.

Intrinsically, in \cref{thr:backward_transfer}, $\mathcal{G}^k_j$ measures the distance between two distributions. By the definition of transition probability in \cref{eq:probability}, $\mathcal G^k_j$ can be rewritten as:
\begin{equation}
\label{eq:bayes}
\begin{aligned}
    \mathcal G^k_j & =\bigl(1-\frac{t^{k\leftarrow{1}}_{y^k j}}{\mathbf p^k[y^k]}\bigr)\mathbf p^1[j] \\
    & = p(Y^1=j|\mathbf x)-\frac{p(Y^k=y^k|Y^1=j,\mathbf x)p(Y^1=j|\mathbf x)}{p(Y^k=y^k)} \\
    & = p(Y^1=j|\mathbf x)-p(Y^1=j|Y^k=y^k, \mathbf x),
\end{aligned}
\end{equation}
where the third equal sign holds due to Bayes' rule. \cref{eq:backward_transfer,eq:bayes} further demonstrate that the $k$-th hierarchy can facilitate logit adjustment for the finest-level one by modifying the distance between $p(Y^1|\mathbf x)$ and $p(Y^1|Y^k=y^k,\mathbf x)$ during the end-to-end training phase. Especially in missing-label scenarios, \cref{eq:backward_transfer,eq:bayes} indicate that the coarser-level hierarchies can provide supervision for the finest-level hierarchy. In fact, the adjustment induced by the $k$-th hierarchy positively impacts the label prediction of the finest-level hierarchy. This can be summarized in the following theorem:

\begin{theorem}[]
\label{thr:distribution_alignment}
In missing-label scenarios where the training sample $\mathbf x$ is partially annotated at the $k$-th hierarchy with ground-truth label $y^k$. If all entries of the feature vector input to the linear classification layer are non-zero \footnote{This assumption is practical, as we can introduce a small positive bias term (e.g, $\epsilon=10^{-8}$) or directly apply a \texttt{sigmoid} activation before the linear classification layer to ensure the input is greater than 0.}, and the network parameters are updated by $\mathcal L^k\triangleq H(\mathbf p^k, \mathbf y^k)$ using SGD-based optimizers, then at the minimum, the following equality holds:
\begin{equation}
    p(Y^1|\mathbf x;\omega) = p(Y^1|Y^k=y^k,\mathbf x;\omega,\theta),
\end{equation}
where $p(Y^1|\mathbf x;\omega)$ is formulated by the Classification Network with parameters $\omega$, and $p(Y^1|Y^k=y^k,\mathbf x;\omega,\theta)$ is jointly formulated by the Classification Network and the Transition Networks with parameters $\omega$ and $\theta=\{\theta^1,\cdots,\theta^k\}$, respectively.
\end{theorem}
We present the proof of \cref{thr:distribution_alignment} in Appendix A.5. Since $p(Y^1|Y^k=y^k,\mathbf{x};\omega,\theta)$ represents the predicted label distribution at the finest-level hierarchy given that the $k$-th hierarchy's prediction is the ground-truth label, this theorem implies that our LHT framework aims to encourage the prediction of the Classification Network $p(Y^1|\mathbf{x};\omega)$ to better match the conditional distribution $p(Y^1|Y^k=y^k,\mathbf{x};\omega,\theta)$. In summary, it concludes that:
\begin{itemize}
    \item The coarse-level hierarchies can provide supervision for the finest-level hierarchy by implicitly minimizing the distance between the two probability distributions $p(Y^1|\mathbf x;\omega)$ and $p(Y^1|Y^k=y^k,\mathbf x;\omega,\theta)$.
    \item The Classification Network tends to predict the labels for the finest-level hierarchy while maintaining the accuracy of the $k$-th hierarchy's prediction during the end-to-end training phase.
\end{itemize}

\subsection{Architectural Designs}
\label{sec:architecture}
As aforementioned, the proposed LHT framework mainly consists of the Classification Network and the Transition Networks. 

Firstly, we emphasize that for the Classification Network as formulated in \cref{eq:fine_level_prediction}, we employ a well-designed existing network, such as ResNet-50 \cite{he2016deep}, to predict category labels for the finest-level hierarchy. On one hand, this aims to insure a fair comparison of classification performance at the finest-level hierarchy with traditional hierarchical classification methods based on the multi-task learning paradigm, while on the other hand, the target is to demonstrate that our proposed LHT is a plug-in for enhancing conventional classification networks through hierarchical supervision, especially in missing-label scenarios.

Next, we elaborate the architectural designs for the Transition Networks. As shown in \cref{fig:framework}(b), for hierarchy $k$, the network  $g_{\theta^k}^k$ is responsible for producing the transition matrix $\mathbf T^k$. A natural way to implement the network is to use a single fully-connected (\texttt{FC}) layer, which can be formulated as:
\begin{equation}
    \label{eq:vanilla_transition_net}
    \mathbf T^k = g^k_{\theta^k}(\mathbf x):=\texttt{softmax}\bigl(\texttt{reshape}(\mathbf W^k\hat{\mathbf x}+\mathbf b^k)\bigr),
\end{equation}
where $\hat{\mathbf x}\in\mathbb{R}^d$ is a high-level embedding of $\mathbf x$ branched from the Classification Network, $\theta^k=\{\mathbf W^k,\mathbf b^k\}$ are the parameters of $g^k_{\theta^k}$, with $\mathbf W^k\in\mathbb{R}^{(C_{k}\times C_{k-1})\times d}$ and $\mathbf b^k\in\mathbb{R}^{(C_{k}\times C_{k-1})}$ denoting the weight matrix and bias term of the \texttt{FC} layer, respectively. The operator $\texttt{reshape}(\cdot)$ rearranges the elements of a $(C_{k}\times C_{k-1})$ vector into a $C_{k}\times C_{k-1}$ matrix, and $\texttt{softmax}(\cdot)$ normalizes the columns of the input matrix using the softmax function for satisfying $\sum_{i=1}^{C_k}\mathbf T_{ij}^k=1$.

From \cref{eq:vanilla_transition_net}, we can observe that the Transition Networks capture sample-specific information through the weight matrix $\mathbf{W}^k$ and sample-agnostic information through the bias term $\mathbf b^k$. This indicates that the learned transition probabilities are conditioned on both the input sample and categorical knowledge. Therefore, the architectural design of the Transition Network complies with \cref{eq:probability}.

\begin{figure}[t]
    \centering
    \includegraphics[width=0.65\linewidth]{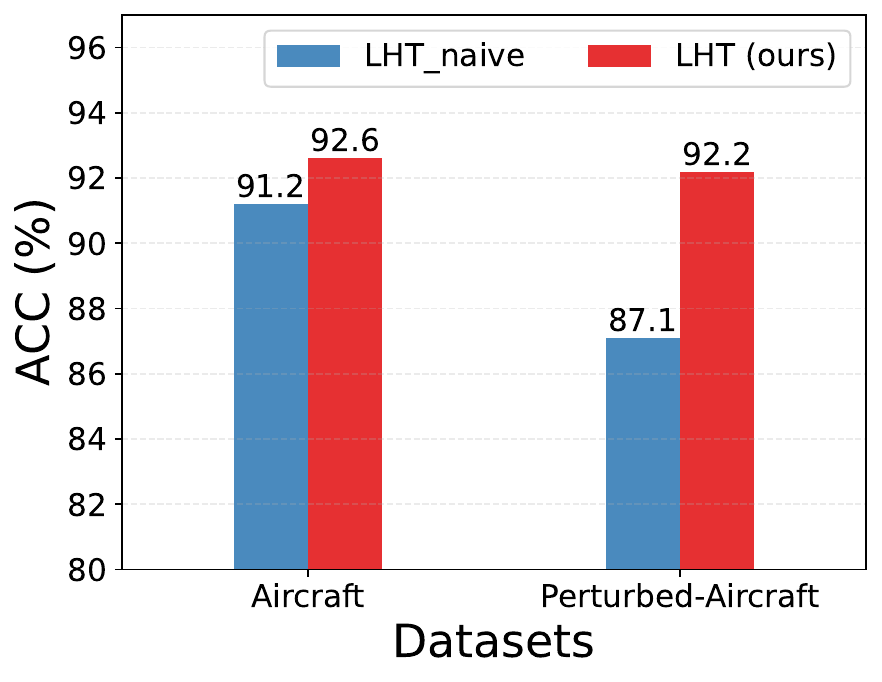}
    \caption{Performance comparison between LHT\_naive and LHT (ours) on the Aircraft and Perturbed-Aircraft datasets.}
    \label{fig:comparison_hist}
\end{figure}

\medskip
\noindent\textbf{Remark 1:} From the given hierarchical labels, we can obtain a deterministic transition matrix $\mathbf S^k$, where only one element is 1 in each column. This indicates that $\mathbf S^k_{i,j}=1$ if category $j$ at hierarchy $k\!-\!1$ is a subclass of category $i$ at hierarchy $k$, and $\mathbf S^k_{i,j}=0$ otherwise. Instead, in Eq. \eqref{eq:vanilla_transition_net}, we aims to model $p(y^k|y^{k-1},\mathbf x)$, which can capture data-driven correlation from image context and hierarchy prior. In fact, if we replace $g_{\theta^k}^k(\hat{\mathbf x})$ with $\mathbf S^k$ in Eq. \eqref{eq:hierarchical_ce_lht} (referred to as LHT\_naive), it will result in a significant performance degradation, as the hierarchical relationships derived from a label tree are not always visually grounded. For example, the Aircraft dataset \cite{maji2013fine}, whose hierarchical relationships are based on a \textit{maker-family-model} organization, is defined more by semantics than by visual similarity. As shown in \cref{fig:comparison_hist}, LTH\_naive achieves notably lower performance than our LHT model. To further demonstrate the importance of adaptively learning visual correlations, we constructed a new dataset, Perturbed-Aircraft, which contains a label hierarchy that is less visually consistent than the original dataset Aircraft \footnote{To construct this dataset based on Aircraft, we retain the original three-level hierarchy but randomly perturb 40\% of the category indices at the second hierarchy, which introduces significant disorder and misalignment in the category relationships compared to Aircraft.}. Under this challenging scenario, our LHT model surpasses LHT\_naive by 5.1\%, further demonstrating the effectiveness of learning data-driven transition matrices to capture visual category correlations.

\begin{figure}[t]
    \centering
    \includegraphics[width=1.0\linewidth]{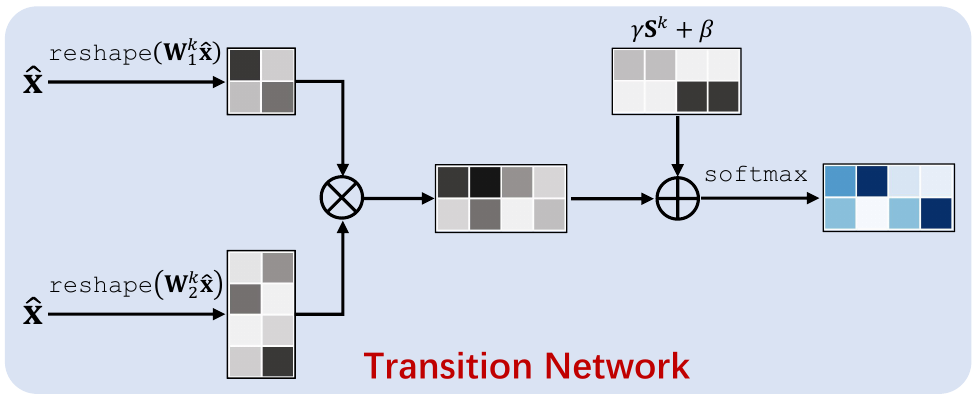}
    \caption{Architecture of the proposed Transition Network. For the $k-$th hierarchy, the network includes three small modules: two smaller \texttt{FC} layers with each followed by a \texttt{reshape} operator, i.e., $\texttt{reshape}(\mathbf W_i^k\hat{\mathbf x}),i=1,2$, and one bias offset term $\gamma S^k+\beta$.}
    \label{fig:transition_network}
\end{figure}

\medskip
\noindent\textbf{A lightweight design for $g_{\theta^k}^k$:} The architecture of $g_{\theta^k}^k$ designed in \cref{eq:vanilla_transition_net} requires a large number of parameters. To prevent an explosion of parameters, we propose a simple design for $g_{\theta^k}^{k}$ as shown in \cref{fig:transition_network}. 

Firstly, we replace the original \texttt{FC} layer with two smaller ones, which means we use a low-rank approximation
to estimate $\mathbf W^k\hat{\mathbf x}$ in \cref{eq:vanilla_transition_net}. This can be formulated as: 
\begin{equation}
    \label{eq:low_rank_approx}
    \texttt{reshape}(\mathbf W^k\hat{\mathbf x}) :\approx \underbrace{\texttt{reshape}(\mathbf W^k_1\hat{\mathbf x})}_{\mathbb R^{C_k\times s}}\otimes\underbrace{\texttt{reshape}(\mathbf W^k_2\hat{\mathbf x})}_{\mathbb R^{C_{k-1}\times s}},
\end{equation}
where $\mathbf W^k_1\in\mathbb R^{(C_k\times s)\times d}$ and $\mathbf W^k_2\in\mathbb R^{(C_{k-1}\times s)\times d}$ are parameters of the two $\texttt{FC}$ layers, with $s$ denoting a hyper-parameter in order to control the rank of the above approximation. The operation $\mathbf A\otimes \mathbf B$ denotes the matrix multiplication between $\mathbf A$ and $\mathbf B^\top$ (i.e., the transpose of $\mathbf B$). \cref{eq:low_rank_approx} indicates that $\mathbf W\hat{\mathbf{x}}$ is approximated with a rank $s$ matrix, which reduces its parameter complexity from $\mathcal{O}\bigl(d(C_k\times C_{k-1})\bigr)$ to $\mathcal{O}\bigl(ds(C_k + C_{k-1})\bigr)$. We initialize $s$ as 2, which ensures efficiency and good performance throughout our experiments. The ablation study on $s$ can be found in \cref{sec:ablation}.

Next, we aim to design a lightweight bias term for replacing $\texttt{reshape}(\mathbf b^k)$ in \cref{eq:vanilla_transition_net} as it involves a relatively large parameter complexity $\mathcal O(C_k\times C_{k-1})$. As previously discussed, this bias term plays a crucial role in capturing sample-agnostic information and encoding categorical knowledge to satisfy \cref{eq:probability}. Interestingly, the deterministic transition matrix $\mathbf S^k$, derived from the given hierarchical labels, can serve as a valuable prior for encoding categorical granularity information. Therefore, we replace $\texttt{reshape}(\mathbf b^k)$ with the following simplified formulation:
\begin{equation}
    \label{eq:bias_term}
    \texttt{reshape}(\mathbf b^k):\approx \gamma \mathbf S^k + \beta,
\end{equation}
where $\gamma$ is a scaling parameter to ensure consistency in magnitude between $\gamma\mathbf S^k$ and $\mathbf W^k\hat{\mathbf x}$, and $\beta$ is a shifting parameter. From \cref{eq:bias_term}, we can see that this new bias term introduces negligible parameters compared to the original version.

By substituting \cref{eq:low_rank_approx,eq:bias_term} into \cref{eq:vanilla_transition_net}, we obtain a lightweight formulation of the Transition Network: $\label{eq:light_transition_net} g_{\theta^k}^k(\mathbf x)=\texttt{softmax}\bigl(\texttt{reshape}(\mathbf W^k\hat{\mathbf x})+\texttt{reshape}(\mathbf b^k)\bigr)$, with the parameters $\theta^k=\{\mathbf W_1^k,\mathbf W_2^k,\gamma,\beta\}$. We emphasize that our basic principle in designing the transition network is to integrate both sample-specific and sample-agnostic information, as investigated in \cref{eq:probability}.

\subsection{Confusion Loss}
\label{sec:confusion_loss}
As aforementioned in \cref{subsec:overall}, the parameters involved in the Classification Network and the Transition Networks, i.e., $w$ and $\theta=\{\theta^k\}_{k=2}^K$, can be jointly optimized end-to-end using the loss \cref{eq:hierarchical_ce_lht}. However, the network training by minimizing this objective function remains the following challenges: (\romannumeral1) The manually annotated labels within each hierarchy are one-hot, and such a cross-entropy loss encourages the network to move all categories toward orthogonality at each hierarchy, counteracting the inter-class similarity introduced by class hierarchies; (\romannumeral2) From the perspective of numerical optimization, there are infinite combinations of $\mathbf T^k$ and ${\mathbf p}^{k-1}$, which easily make the learned transition matrices to collapse to a trivial solution.

\begin{algorithm}[t]
   \caption{Mini-batch training algorithm of LHT}
   \label{algorithm}
   {\bfseries Input:} Training data $\mathcal D$, hierarchy level $K$, batch size $N$, training iteration $T$, learning rate $\alpha$. \\
   {\bfseries Output:} Parameters of the Classification Network and Transition Networks: $\omega$, $\theta=\{\theta^k\}_{k=2}^K$.
\begin{algorithmic}[1]
   \STATE Initialize the parameters $\omega^{(1)}$ and $\theta^{(1)}$.
   \FOR{$t=1$ {\bfseries to} $T$}
   \STATE Sample a mini-batch of training samples from $\mathcal D$: \\ $\hat{\mathcal D}=\{\mathbf x_n, (y_n^1,\cdots,y_n^K)\}_{n=1}^N$.
        \FOR {$k=1$ {\bfseries to} $K$}
        \IF{k=1}
        \STATE Compute $\mathbf p^1_n$ by Eq. \eqref{eq:fine_level_prediction}.
        \ELSE
        \STATE Compute $\mathbf p^k_n$ by Eq. \eqref{eq:coarse_level_prediction}.
        \ENDIF
        \ENDFOR
   \STATE Compute the classification loss $\mathcal L_{\text{CE}}$ by Eq. \eqref{eq:hierarchical_ce_lht}.
   \STATE Compute the confusion loss $\mathcal L_{\text{Conf}}$ by Eq. \eqref{eq:loss_conf}.
   \STATE Compute the overall training loss $\mathcal L$ by Eq. \eqref{eq:loss}.
   \STATE Update $\{\omega^{(t+1)}, \theta^{(t+1)}\}=\{\omega^{(t+1)}, \theta^{(t+1)}\}-\alpha\nabla_{\omega,\theta}\mathcal L$.
   \ENDFOR
\end{algorithmic}
    {\bfseries Return:} Optimized parameters $\omega^{(T)}$ and $\theta^{(T)}$
\end{algorithm}

To address the aforementioned challenges, we further introduce an additional regularization loss on transition matrices $\{\mathbf T^k\}_{k=2}^K$, which we call the confusion loss $\mathcal L_{{\rm Conf}}$. We herein revisit the label hierarchy transition matrices involved in our LHT model. The column vector (e.g., the $j$-th one) of transition matrix $\mathbf T^k$ represents the label distribution of the $k$-th hierarchy when the sample is labeled as class $j$ at the $(k\!-\!1)$-th hierarchy. Each element of this vector denotes the likelihood of similarity, or confusion, between two classes across two adjacent hierarchies. The proposed confusion loss aims to encourage such class confusion to counter the over-confident prediction induced by the cross-entropy loss. Mathematically, for a batch of $N$ training data $\{\mathbf x_n, (y_n^1, y_n^2,\cdots,y_n^K)\}_{n=1}^N$, the confusion loss is defined as the negative entropy of the column space spanned by the transition matrices, i.e.,
\begin{equation}
    \mathcal L_{\rm Conf}=\sum_{n=1}^N\sum_{k=2}^K\sum_{j=1}^{C_{k-1}}\frac{1}{C_{k-1}} \Bigl\langle\mathbf T_n^k[:,j],\log
    \mathbf T_n^k[:,j]\Bigr\rangle,
    \label{eq:loss_conf}
\end{equation}
where $\mathbf T_n^k$ is the predicted transition matrix for $\mathbf x_n$ and $\mathbf T_n^k[:,j]$ denotes its $j$-th column, $\log(\cdot)$ is element-wise logarithm operation for vectors, and $\langle \cdot, \cdot\rangle$ is the inner product operation of vectors. Through minimization of Eq. \eqref{eq:loss_conf}, the conditional distribution represented by each column of $\mathbf T_n^k$ is encouraged to have a higher entropy (i.e., this distribution inclines to approach uniform distribution). In other words, this introduces \emph{confusion} in fine-to-coarse label transitions during training, thereby encouraging the Transition Networks to encode sample-specific information into transition matrices and alleviating the over-confident prediction induced by the hierarchical cross-entropy loss. Furthermore, since $\mathbf T^{k\leftarrow 1}$ can be factorized as $\mathbf T^{k\leftarrow 1}=\mathbf T^k \cdots \mathbf T^2$ (Proposition~1), applying confusion loss to each transition matrix may implicitly facilitate the modeling of inter-class similarities between hierarchy 1 and $k$.

Combined Eq. \eqref{eq:hierarchical_ce_lht} and Eq. \eqref{eq:loss_conf}, the overall training loss can be formulated as
\begin{equation}
    \mathcal{L}=\mathcal{L}_{{\rm CE}}+\lambda \mathcal L_{{\rm Conf}},
    \label{eq:loss}
\end{equation}
where $\lambda$ is hyper-parameter for trade-off between the cross-entropy term $\mathcal L_{{\rm CE}}$ and the confusion  regularization term $\mathcal L_{{\rm Conf}}$. All the parameters involved in our LHT can be trained in an end-to-end manner. The training algorithm is summarized in Algorithm \ref{algorithm}.

\section{Experiments on Natural Images}
\label{sec:experiment}
In this section, we first demonstrate the effectiveness of the proposed LHT in \textbf{supervised/semi-supervised hierarchical classification} tasks following \cite{chen2022label}. Specifically, the labels of training samples are given at all levels of the class hierarchy under the supervised setting. However, under the semi-supervised setting, the labels are observed at different levels of the class hierarchy, indicating that part of samples are only observed at internal levels of the whole hierarchies\footnote{In this paper, we stem the term of “semi-supervised learning” from \cite{chen2022label}, in the sense of ``partially supervised with incomplete hierarchical labels for a part of training samples", in contrast to the classical term that assumes most training samples are given without any annotation.}. Next, we demonstrate the applicability of the proposed LHT on \textbf{large-scale low-shot learning} where transferable visual features are learned from a class hierarchy to encode the semantic relations between source and target classes. We also provide thorough ablation studies to gain insight into our method by showing the efficacy of each key component.
\subsection{Supervised/Semi-supervised Hierarchical Learning}

\subsubsection{Experimental Setups}

\textbf{Datasets.} In our experiments, we evaluate our method on three public benchmark datasets: CUB-200-2011 \cite{wah2011caltech},  Aircraft \cite{maji2013fine} and Stanford Cars \cite{krause20133d}, where the label hierarchies are publicly available or constructed by Chang et al. \cite{chang2021your} according to their lexical relationships in Wikipedia. The details of these datasets are as follows:

\begin{itemize}
    \item \textbf{CUB-200-2011}\footnote{\href{https://www.vision.caltech.edu/datasets/cub_200_2011}{http://www.vision.caltech.edu/visipedia/CUB-200-2011.html}} is a dataset of bird images mainly introduced for the problem of fine-grained visual categorisation methods. It contains 11,877 images belonging to 200 bird species. For hierarchical classification evaluation, the labels are re-organized into three-level hierarchy with 13 orders, 38 families and 200 species by tracing their biological taxonomy \cite{chang2021your}. For example, the original label ``\emph{Brewer Blackbird}" is spanned into (``\emph{Passeriformes}'', ``\emph{Icteridae}'', ``\emph{Brewer Blackbird}'') in the order of (Order, Family, Species).
    \item\textbf{Aircraft}\footnote{\href{https://www.robots.ox.ac.uk/~vgg/data/fgvc-aircraft}{https://www.robots.ox.ac.uk/~vgg/data/fgvc-aircraft}} consists of 10,000 images across 100 airplane classes, with a data ratio of approximately 2:1 between the training and test sets. The dataset includes a three-level label hierarchy with 30 makers (e.g., ``\emph{Boeing}''), 70 families (e.g., ``\emph{Boeing 767}'') and 100 models (e.g., ``\emph{767-200}''). As a result, this dataset can be readily used in our experiments without any modification.

    \item \textbf{Stanford Cars}\footnote{\href{https://www.kaggle.com/datasets/jessicali9530/stanford-cars-dataset}{https://www.kaggle.com/datasets/jessicali9530/stanford-cars-dataset}} contains 8,144 training and 8,041 test images across 196 car classes. It is re-organized into two-level label hierarchy, i.e., 9 car types and 196 car classes as introduced in \cite{chang2021your}. For example, the class label ``\emph{BMW X6 SUV 2012}'' are spanned into (``\emph{SUV}'', ``\emph{BMW X6 SUV 2012}''). We use the standard public train/test splits, and not any bounding box/part annotations are involved in all our experiments.
\end{itemize}

For the supervised hierarchical classification setting, we train the model using the complete set of labels at all hierarchical levels for each training sample. For the semi-supervised hierarchical classification setting, in order to simulate a lack of domain knowledge, we randomly select 30\%, 50\%, 70\%, and 90\% samples (denoted as $\mathcal R=90\%$ for example) to mask out their labels at the finest-level hierarchy, such that their intermediate parent classes are relabeled as the leaf classes. When employing this approach in semi-supervised scenarios, in our model training process, we simply neglect the loss imposed on the finest-level hierarchy for the samples that have been relabeled to their parent classes.

\begin{table*}[t]
  \renewcommand\arraystretch{1.1}
  \centering
  \caption{Comparison results of all competing methods over \emph{ACC} and \emph{mACC} on CUB-200-2011 dataset. $R$ denotes that a fraction of $R\%$ of the training samples lack labels at the finest-level (Species) hierarchy. The best results are highlighted in \textbf{bold}. $^*$ indicates the results reported in the original paper.}
  \label{tab:sota_cub}
  \setlength{\tabcolsep}{1.0mm}{
  \begin{tabular}{c c cc cc cc cc cc cc cc cc}
    \toprule
    \multirow{2}{*}{$R$} & \multirow{2}{*}{Hierarchy} & \multicolumn{2}{c}{HD-CNN~\cite{yan2015hd}} & \multicolumn{2}{c}{HMCN~\cite{wehrmann2018hierarchical}} & \multicolumn{2}{c}{C-HMCNN~\cite{giunchiglia2020coherent}} & 
    \multicolumn{2}{c}{\revised{Soft Labels}~\cite{bertinetto2020making}} &
    \multicolumn{2}{c}{FGoN~\cite{chang2021your}} & \multicolumn{2}{c}{HRN\cite{chen2022label}} & \multicolumn{2}{c}{HRN\cite{chen2022label}$^*$} & \multicolumn{2}{c}{Ours} \\
    \cmidrule(r){3-4} \cmidrule(r){5-6} \cmidrule(r){7-8} \cmidrule(r){9-10} \cmidrule(r){11-12} \cmidrule(r){13-14} \cmidrule(r){15-16} \cmidrule(r){17-18}
    ~ & ~ & ACC & mACC & ACC & mACC & ACC & mACC & \revised{ACC} & \revised{mACC} & ACC & mACC  & ACC & mACC & ACC & mACC & ACC & mACC \\
    \midrule
    \multirow{3}{*}{0\%} & Order & 98.5 & \multirow{3}{*}{93.3} & 98.2 & \multirow{3}{*}{93.0} & 98.6 & \multirow{3}{*}{93.6} &\revised{98.8} & \multirow{3}{*}{\revised{93.7}} & 98.8 & \multirow{3}{*}{93.6} & 98.8 & \multirow{3}{*}{93.7} & 98.7 & \multirow{3}{*}{93.6} & \textbf{99.0} & \multirow{3}{*}{\textbf{94.1}} \\
                         & Family & 95.5 &  & 94.7 &  & 95.7 &  & \revised{95.9} & & 95.5 &  & 95.5 &  & 95.5 &  & \textbf{96.2} \\
                         & Species & 85.9 &  & 86.0 &  & 86.5 & & \revised{86.5} & & 86.6 &  & 86.7 &  & 86.6 &  & \textbf{87.0}\\
\hline
    \multirow{3}{*}{30\%} & Order & 98.5 & \multirow{3}{*}{92.5} & 98.0 & \multirow{3}{*}{91.9} & 98.5 & \multirow{3}{*}{92.8} & \revised{98.9} & \multirow{3}{*}{\revised{93.1}} & 98.7 & \multirow{3}{*}{93.0} & 98.5 & \multirow{3}{*}{92.4} & 98.3 & \multirow{3}{*}{92.3} & \textbf{98.9} & \multirow{3}{*}{\textbf{93.7}} \\
                         & Family & 95.3 &  & 94.0 &  & 95.2 &  & \revised{95.7} &  & 95.2 &  & 95.0 &  & 94.8 &  & \textbf{96.2}\\
                         & Species & 83.8 &  & 83.6 &  & 84.6 &  & \revised{84.8} &  & 85.1 &  & 83.6 &  & 83.9 &  & \textbf{86.0} \\
\hline
    \multirow{3}{*}{50\%} & Order & 98.4 & \multirow{3}{*}{91.7} & 97.9 & \multirow{3}{*}{90.2} & 98.5 & \multirow{3}{*}{91.9} & \revised{98.6} & \multirow{3}{*}{\revised{92.1}} & 98.7 & \multirow{3}{*}{92.3} & 98.0 & \multirow{3}{*}{91.3} & 97.9 & \multirow{3}{*}{90.9} & \textbf{98.9} & \multirow{3}{*}{\textbf{93.3}} \\
                         & Family & 94.7 &  & 93.5 &  & 94.8 &  & \revised{95.0} &  & 95.3 &  & 94.7 &  & 94.3 &  & \textbf{96.1} \\
                         & Species & 81.9 &  & 79.2 &  & 82.3 &  & \revised{82.7} &  & 82.8 &  & 81.1 &  & 80.5 &  & \textbf{84.8} \\
\hline
    \multirow{3}{*}{70\%} & Order & 98.4 & \multirow{3}{*}{90.1} & 97.7 & \multirow{3}{*}{87.1} & 96.9 & \multirow{3}{*}{87.8} & \revised{97.6} & \multirow{3}{*}{\revised{89.3}} & 98.7 & \multirow{3}{*}{90.7} & 97.2 & \multirow{3}{*}{88.3} & 98.4 & \multirow{3}{*}{88.8} & \textbf{98.9} & \multirow{3}{*}{\textbf{92.4}} \\
                         & Family & 94.7 &  & 92.7 &  & 91.3 &  & \revised{92.6} &  & 94.9 &  & 93.3 &  & 93.9 &  & \textbf{96.0}\\
                         & Species & 77.2 &  & 70.9 &  & 75.2 &  & \revised{77.8} &  & 78.6 &  & 74.4 &  & 74.0 &  & \textbf{82.4} \\
\hline
    \multirow{3}{*}{90\%} & Order & 98.5 & \multirow{3}{*}{83.8} & 97.8 & \multirow{3}{*}{79.3} & 91.2 & \multirow{3}{*}{72.6} & \revised{94.1} & \multirow{3}{*}{\revised{79.1}} & 98.7 & \multirow{3}{*}{85.6} & 97.7 & \multirow{3}{*}{83.4} & 98.0 & \multirow{3}{*}{81.4} & \textbf{98.8} & \multirow{3}{*}{\textbf{87.6}} \\
                         & Family & 94.6 &  & 90.6 &  & 77.1 &  & \revised{83.4} &  & 94.3 &  & 93.6 &  & 93.3 &  & \textbf{95.3} \\
                         & Species & 58.3 &  & 49.4 &  & 49.6 &  & \revised{59.7} &  & 63.8 &  & 58.8 &  & 53.0 &  & \textbf{68.8} \\
    \bottomrule
  \end{tabular}
  }
\end{table*}

\begin{table*}[t]
  \renewcommand\arraystretch{1.1}
  \centering
  \caption{Comparison results of all competing methods over \emph{ACC} and \emph{mACC} on Aircraft dataset. $R$ denotes that a fraction of $R\%$ of the training samples lack labels at the finest-level (Model) hierarchy. The best results are highlighted in \textbf{bold}. $^*$ indicates the results reported in the original paper.}
  \label{tab:sota_air}
  \setlength{\tabcolsep}{1.0mm}{
  \begin{tabular}{c c cc cc cc cc cc cc cc cc}
    \toprule
    \multirow{2}{*}{$R$} & \multirow{2}{*}{Hierarchy} & \multicolumn{2}{c}{HD-CNN~\cite{yan2015hd}} & \multicolumn{2}{c}{HMCN~\cite{wehrmann2018hierarchical}} & \multicolumn{2}{c}{C-HMCNN~\cite{giunchiglia2020coherent}} & 
    \multicolumn{2}{c}{\revised{Soft Labels}~\cite{bertinetto2020making}} &
    \multicolumn{2}{c}{FGoN~\cite{chang2021your}} & \multicolumn{2}{c}{HRN\cite{chen2022label}} & \multicolumn{2}{c}{HRN\cite{chen2022label}$^*$} & \multicolumn{2}{c}{Ours} \\
    \cmidrule(r){3-4} \cmidrule(r){5-6} \cmidrule(r){7-8} \cmidrule(r){9-10} \cmidrule(r){11-12} \cmidrule(r){13-14} \cmidrule(r){15-16} \cmidrule(r){17-18}
    ~ & ~ & ACC & mACC & ACC & mACC & ACC & mACC & \revised{ACC} & \revised{mACC} & ACC & mACC  & ACC & mACC & ACC & mACC & ACC & mACC \\
    \midrule
    \multirow{3}{*}{0\%} & Maker & 96.9 & \multirow{3}{*}{94.3} & 97.2 & \multirow{3}{*}{95.1} & 97.1 & \multirow{3}{*}{94.6} & \revised{97.1} & \multirow{3}{*}{\revised{94.4}} & 97.1 & \multirow{3}{*}{94.7} & 97.3 & \multirow{3}{*}{95.2} & \textbf{97.5} & \multirow{3}{*}{\textbf{95.3}} & 97.4 & \multirow{3}{*}{\textbf{95.3}} \\
                         & Family & 94.9 &  & 95.6 &  & 95.4 &  & \revised{95.0} &  & 95.5 &  & 95.7 &  & 95.8 &  & \textbf{95.9} \\
                         & Model & 91.1 &  & 92.6 &  & 91.4 &  & \revised{91.1} &  & 91.4 &  & \textbf{92.7} &  & 92.6 &  & 92.6 \\
\hline
    \multirow{3}{*}{30\%} & Maker & 96.9 & \multirow{3}{*}{93.7} & 97.2 & \multirow{3}{*}{94.8} & 96.3 & \multirow{3}{*}{93.1} & \revised{96.8} & \multirow{3}{*}{\revised{93.3}} & 96.9 & \multirow{3}{*}{94.1} & \textbf{97.3} & \multirow{3}{*}{94.5} & \textbf{97.3} & \multirow{3}{*}{94.8} & 97.2 & \multirow{3}{*}{\textbf{95.0}} \\
                         & Family & 94.6 &  & 95.6 &  & 94.0 &  & \revised{94.1} &  & 95.4 &  & 95.3 &  & 95.5 &  & \textbf{95.8} \\
                         & Model & 89.7 &  & 91.6 &  & 89.0 &  & \revised{89.1} &  & 90.0 &  & 91.0 &  & 91.6 & & \textbf{91.9} \\
\hline
    \multirow{3}{*}{50\%} & Maker & 96.8 & \multirow{3}{*}{93.4} & 96.9 & \multirow{3}{*}{93.8} & 95.9 & \multirow{3}{*}{92.1} & \revised{96.0} & \multirow{3}{*}{\revised{91.9}} & 96.9 & \multirow{3}{*}{93.5} & 96.8 & \multirow{3}{*}{94.0} & \textbf{97.3} & \multirow{3}{*}{94.2} & \textbf{97.3} & \multirow{3}{*}{\textbf{94.8}} \\
                         & Family & 94.8 &  & 95.1 &  & 93.2 &  & \revised{93.0} &  & 95.3 &  & 95.1 &  & 95.7 &  & \textbf{95.8} \\
                         & Model & 88.7 &  & 89.3 &  & 87.1 &  & \revised{86.6} &  & 88.2 &  & 90.1 &  & 89.7 & & \textbf{91.4} \\
\hline
    \multirow{3}{*}{70\%} & Maker & 96.7 & \multirow{3}{*}{92.3} & 96.9 & \multirow{3}{*}{92.1} & 93.4 & \multirow{3}{*}{88.0} & \revised{94.1} & \multirow{3}{*}{\revised{88.0}} & 96.9 & \multirow{3}{*}{92.0} & 96.2 & \multirow{3}{*}{91.9} & 96.8 & \multirow{3}{*}{91.8} & \textbf{97.3} & \multirow{3}{*}{\textbf{93.6}}\\
                         & Family & 94.5 &  & 94.4 &  & 89.7 &  & \revised{89.8} &  & 95.2 &  & 93.8 &  & 94.2 &  & \textbf{95.3}\\
                         & Model & 85.8 &  & 84.9 &  & 80.8 &  & \revised{80.0} &  & 83.8 &  & 85.7 &  & 84.5 &  & \textbf{88.1} \\
\hline
    \multirow{3}{*}{90\%} & Maker & 96.6 & \multirow{3}{*}{89.0} & 96.9 & \multirow{3}{*}{89.1} & 73.7 & \multirow{3}{*}{61.6} & \revised{82.4} & \multirow{3}{*}{\revised{71.0}} & 96.4 & \multirow{3}{*}{87.6} & 96.3 & \multirow{3}{*}{87.3} & 95.4 & \multirow{3}{*}{86.1} & \textbf{97.1} & \multirow{3}{*}{\textbf{90.5}} \\
                         & Family & 93.8 &  & 94.7 &  & 63.0 &  & \revised{72.4} &  & 94.6 &  & 92.5 &  & 91.7 &  & \textbf{95.0} \\
                         & Model & 76.7 &  & 75.8 &  & 48.1 &  & \revised{58.2} &  & 71.9 &  & 73.2 &  & 71.1 & & \textbf{79.3} \\
    \bottomrule
  \end{tabular}
  }
\end{table*}

\begin{table*}[t]
  \renewcommand\arraystretch{1.1}
  \centering
  \caption{Comparison results of all competing methods over \emph{ACC} and \emph{mACC} on Stanford Cars dataset. $R$ denotes that a fraction of $R\%$ of the training samples lack labels at the finest-level (Maker) hierarchy. The best results are highlighted in \textbf{bold}. $^*$ indicates the results reported in the original paper.}
  \label{tab:sota_car}
  \setlength{\tabcolsep}{1.0mm}{
  \begin{tabular}{c c cc cc cc cc cc cc cc cc}
    \toprule
    \multirow{2}{*}{$R$} & \multirow{2}{*}{Hierarchy} & \multicolumn{2}{c}{HD-CNN~\cite{yan2015hd}} & \multicolumn{2}{c}{HMCN~\cite{wehrmann2018hierarchical}} & \multicolumn{2}{c}{C-HMCNN~\cite{giunchiglia2020coherent}} & 
    \multicolumn{2}{c}{\revised{Soft Labels}~\cite{bertinetto2020making}} &
    \multicolumn{2}{c}{FGoN~\cite{chang2021your}} & \multicolumn{2}{c}{HRN\cite{chen2022label}} & \multicolumn{2}{c}{HRN\cite{chen2022label}$^*$} & \multicolumn{2}{c}{Ours} \\
    \cmidrule(r){3-4} \cmidrule(r){5-6} \cmidrule(r){7-8} \cmidrule(r){9-10} \cmidrule(r){11-12} \cmidrule(r){13-14} \cmidrule(r){15-16} \cmidrule(r){17-18}
    ~ & ~ & ACC & mACC & ACC & mACC & ACC & mACC & \revised{ACC} & \revised{mACC} & ACC & mACC  & ACC & mACC & ACC & mACC & ACC & mACC \\
    \midrule
    \multirow{2}{*}{0\%} & Type & 96.8 & \multirow{2}{*}{94.9} & 97.0 & \multirow{2}{*}{95.2} & \textbf{97.7} & \multirow{2}{*}{95.4} & \revised{97.2} & \multirow{2}{*}{\revised{95.4}} & 96.7 & \multirow{2}{*}{95.1} & 97.2 & \multirow{2}{*}{95.4} & 97.4 & \multirow{2}{*}{\textbf{95.7}} & 97.0 & \multirow{2}{*}{95.6}\\
                         & Maker & 93.0 &  & 93.4 &  & 93.1 &  & \revised{93.5} &  & 93.4 &  & 93.6 &  & 94.0 & & \textbf{94.1}\\
\hline
    \multirow{2}{*}{30\%} & Type & 96.5 & \multirow{2}{*}{94.2} & 96.6 & \multirow{2}{*}{94.0} & 96.9 & \multirow{2}{*}{93.9} & \revised{97.1} & \multirow{2}{*}{\revised{94.6}} & 96.5 & \multirow{2}{*}{94.2}  & 96.5 & \multirow{2}{*}{94.0} & 96.1 & \multirow{2}{*}{93.4} & \textbf{97.0} & \multirow{2}{*}{\textbf{95.2}} \\
                         & Maker & 91.8 &  & 91.4 &  & 90.9 &  & \revised{92.0} &  & 91.8 &  & 91.4 &  & 90.6 & & \textbf{93.4}\\
\hline
    \multirow{2}{*}{50\%} & Type & 96.3 & \multirow{2}{*}{93.0} & 96.6 & \multirow{2}{*}{92.1} & 95.5 & \multirow{2}{*}{92.0} & \revised{96.1} & \multirow{2}{*}{\revised{93.0}} & 96.2 & \multirow{2}{*}{93.1}  & 96.0 & \multirow{2}{*}{92.2} & 95.9 & \multirow{2}{*}{92.3} & \textbf{96.9} & \multirow{2}{*}{\textbf{94.8}} \\
                         & Maker & 89.7 &  & 87.6 &  & 88.4 &  & \revised{89.8} &  & 90.0 &  & 88.3 &  & 88.7 & & \textbf{92.7}\\
\hline
    \multirow{2}{*}{70\%} & Type & 96.1 & \multirow{2}{*}{90.5} & 96.2 & \multirow{2}{*}{86.8} & 93.4 & \multirow{2}{*}{86.7} & \revised{93.9} & \multirow{2}{*}{\revised{88.8}} & 95.8 & \multirow{2}{*}{89.9}  & 96.0 & \multirow{2}{*}{87.9} & 96.1 & \multirow{2}{*}{89.9} & \textbf{96.9} & \multirow{2}{*}{\textbf{93.4}} \\
                         & Maker & 84.8 &  & 77.3 &  & 80.0 &  & \revised{83.7} &   & 84.0 &  & 79.7 &  & 83.7 & & \textbf{89.9} \\
\hline
    \multirow{2}{*}{90\%} & Type & 95.3 & \multirow{2}{*}{76.6} & 95.2 & \multirow{2}{*}{69.8} & 82.4 & \multirow{2}{*}{65.5} & \revised{80.0} & \multirow{2}{*}{\revised{67.6}} & 95.0 & \multirow{2}{*}{77.4}  & 94.1 & \multirow{2}{*}{71.0} & 94.3 & \multirow{2}{*}{71.8} & \textbf{96.0} & \multirow{2}{*}{\textbf{83.7}} \\
                         & Maker & 57.8 &  & 44.4 &  & 48.5 &  & \revised{55.1} &   & 59.7 &  & 47.9 &  & 49.3 & & \textbf{71.4} \\
    \bottomrule
  \end{tabular}
  }
\end{table*}

\medskip
\noindent\textbf{Evaluation Metrics:} We adopt the same evaluation metrics as Chang et al. \cite{chang2021your} to evaluate our method. Specifically, \emph{ACC} (i.e., accuracy, the percentage of images whose labels are correctly classified) is reported at each class hierarchy, and then \emph{mACC} (the mean of \emph{ACC} across all class hierarchies) is calculated to measure the classification performance at a hierarchical level. Note that, unless otherwise specified, the \emph{ACC} results are reported as the mean over three random seeds of each experiment.

\medskip
\noindent\textbf{Implementation Details:} All our experiments are implemented with the Pytorch platform \cite{paszke2019pytorch} running on NVIDIA Gefore RTX 3090Ti GPUs. Following \cite{chang2021your}, we employ ResNet-50 pre-trained on ImageNet \cite{deng2009imagenet} as our network backbone, followed by a fully-connected layer with 600 hidden units. For fair comparison, the features of fully-connected layer are uniformly split into $K$ parts ($K=2$ for Standford Cars and $K=3$ for other datasets), with each used for predicting the labels at each hierarchy. During training, the models are trained with Momentum SGD with a momentum of 0.9, a weight decay of $5\times10^{-4}$, a batch-size of 8, and an initial learning rate of 0.0002 for backbone layers and 0.002 for other layers. The training epoch is set to be 200 and the learning rate is adjusted by the cosine annealing schedule. We initially set $\lambda=0.1$ for Cub-200-2011 and $\lambda=2.0$ for the other two datasets, except for $\lambda=0.01$ for all the three datasets under a relabeled ratio of $\mathcal R=90\%$. The same augmentation strategies as described in \cite{chen2022label} are adopted, i.e., each image is resized to $448\times448$, and then the resulting image is randomly cropped (random cropping for training and center cropping for testing) and randomly horizontally flipped.

\begin{figure*}[t]
\centering
    \begin{minipage}[t]{0.32\linewidth}
    \centering
    \includegraphics[width=0.95\linewidth]{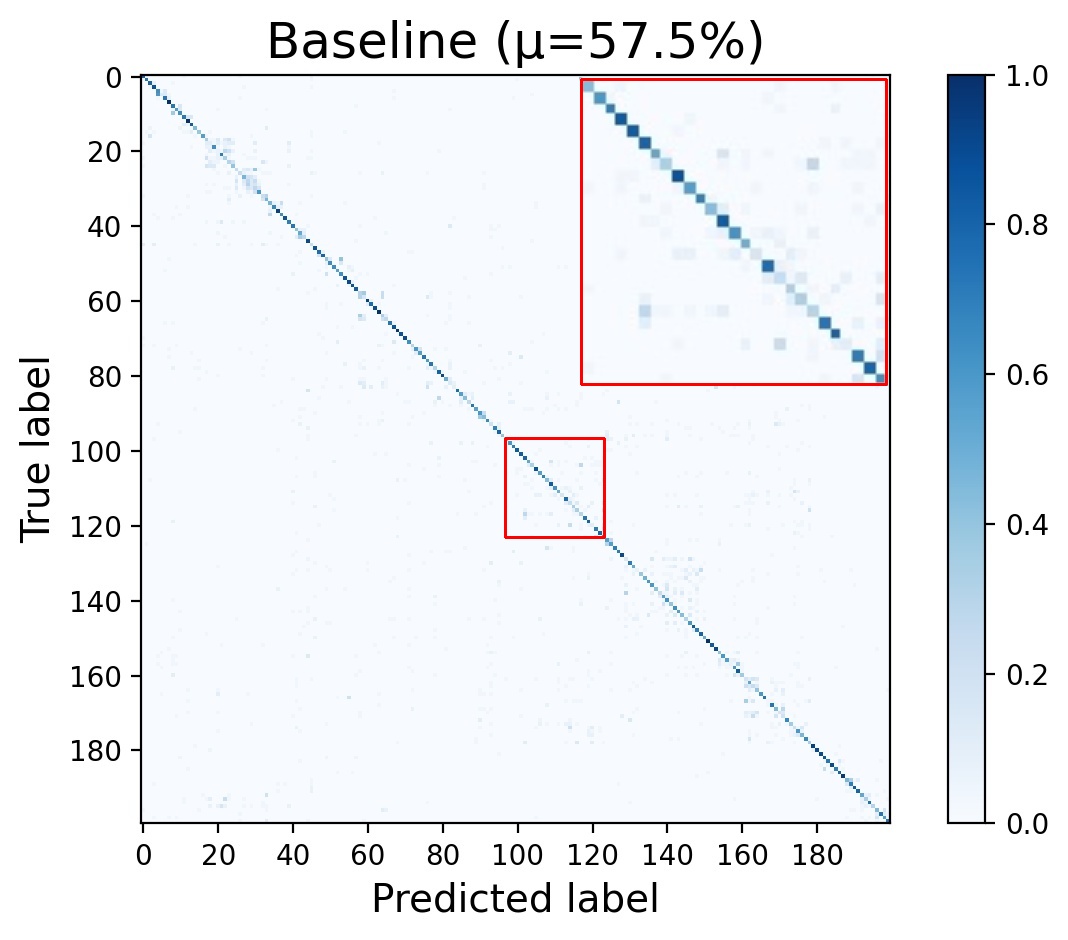}
    \end{minipage}%
    \hfill
    \begin{minipage}[t]{0.32\linewidth}
    \centering
    \includegraphics[width=0.95\linewidth]{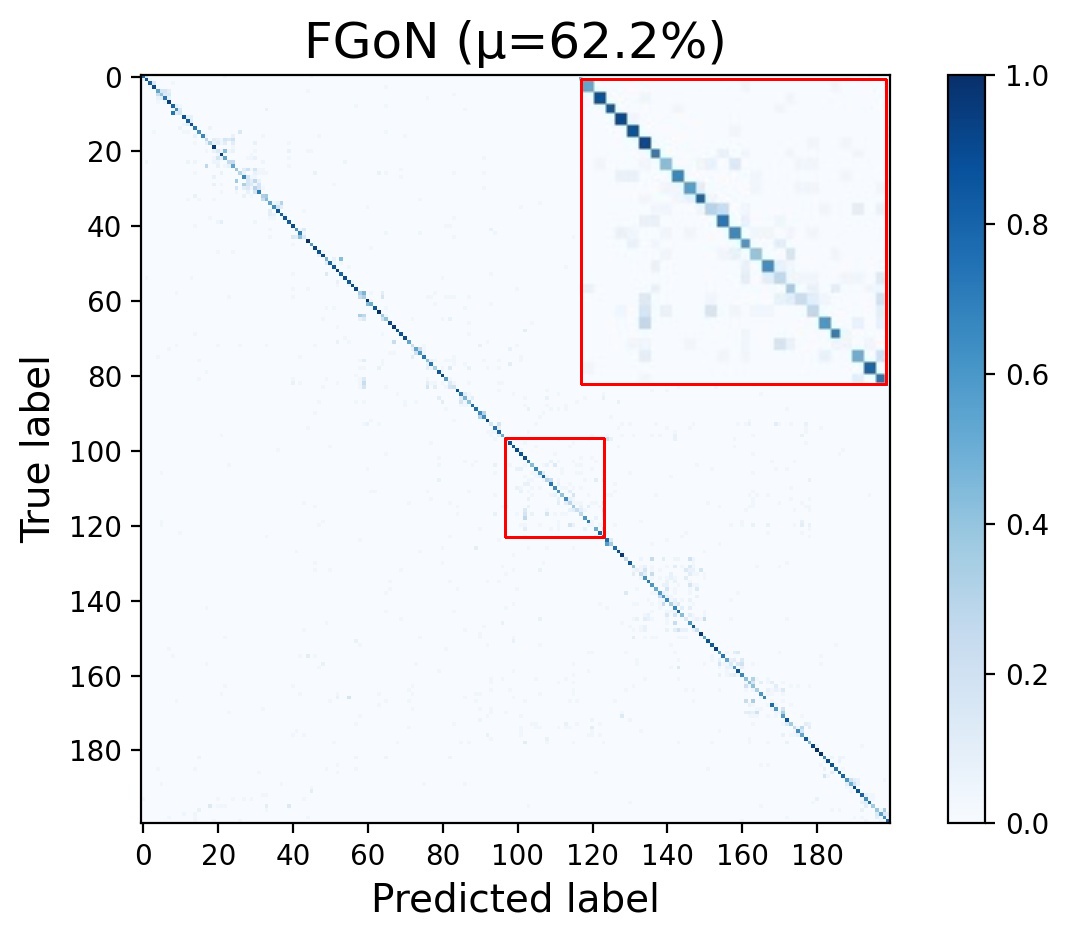}
    \end{minipage}%
    \hfill
    \begin{minipage}[t]{0.32\linewidth}
    \centering
    \includegraphics[width=0.95\linewidth]{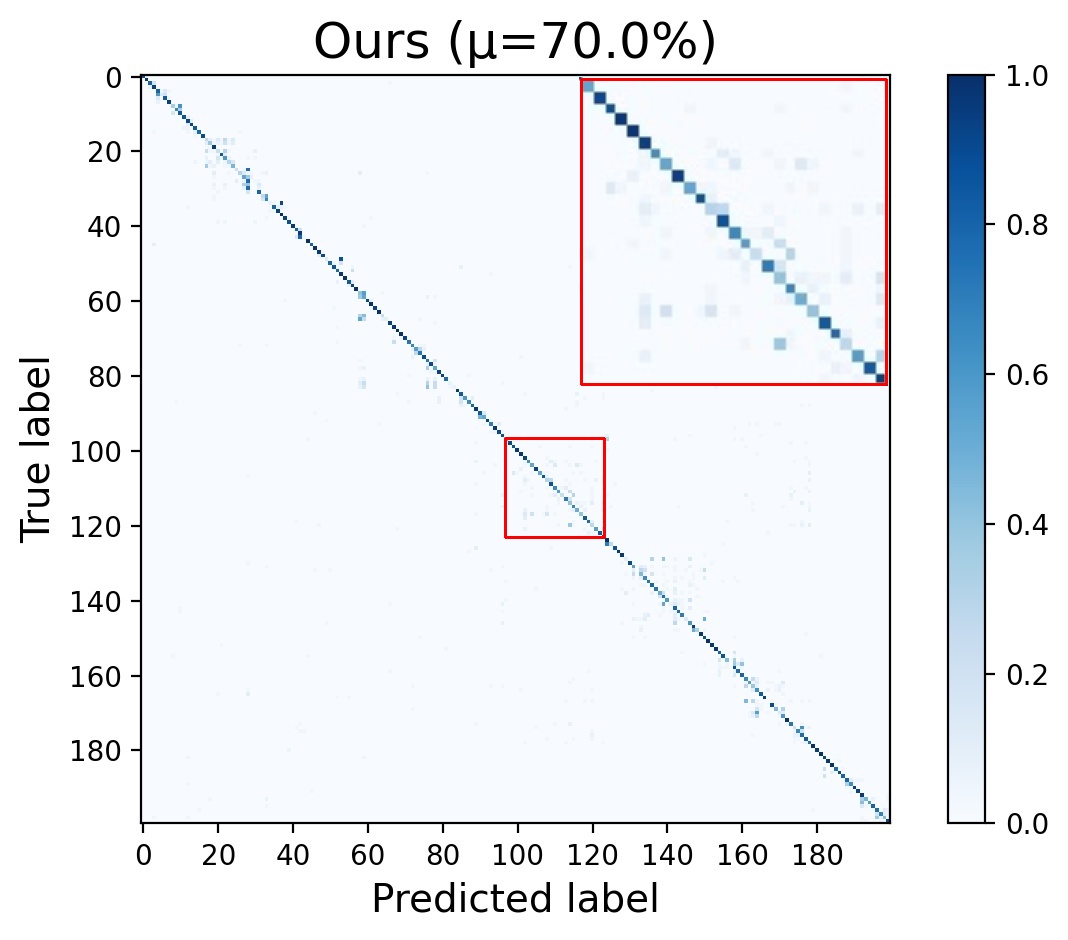}
    \end{minipage}
\caption{Confusion matrices of Baseline, FGoN \cite{chang2021your} and our proposed LHT method (Ours) at the finest-grained hierarchy on Cub-200-2011 with the relabeled ratio $\mathcal R=90$. Here $\mu$ above each figure represents mean-class-recall for all 200 classes.}
\label{fig:confusion}
\end{figure*}

\subsubsection{Main Results}
\label{sec:comparision}
We compare the proposed method with state-of-the-art methods on this task, including: (\romannumeral1) \textbf{HD-CNN}~\cite{yan2015hd}, which aims to learn a hierarchical deep convolutional neural network where a single network backbone was shared by multiple fully-connected layers with each one
responsible for the label prediction within one hierarchy;  (\romannumeral2) \textbf{HMCN}~\cite{wehrmann2018hierarchical}, which designs a specific network architecture with multiple local outputs for label prediction per hierarchical level and one global output layer for capturing semantic correlation in the hierarchy as a whole; (\romannumeral3) \textbf{C-HMCNN}~\cite{giunchiglia2020coherent}, which exploits the hierarchy information by imposing the parent-child constraint on the output of the coarser-level hierarchies, so as to ensure that no hierarchy violation happens (i.e., when the network predicts a sample belonging to a class, this sample also belongs to its parent classes); (\romannumeral4) \textbf{Soft Labels}~\cite{bertinetto2020making}, an embedding-based method that softens the finest-level labels according to the metric derived by the label tree during training, and during testing, we backtrack through the label tree to obtain other coarse-level labels; (\romannumeral5) \textbf{FGoN}~\cite{chang2021your}, one of the current state-of-the-arts which equips with hierarchy-specific classification head to disentangle and reinforce the features of different hierarchies, that is, only forces finer-level features to participate in coarser-level label predictions; (\romannumeral6) \textbf{HRN}~\cite{chen2022label}, which incorporates hierarchical feature interaction into model design and proposes a hierarchical residual network, where hierarchy-specific features from parent level are added to features of children levels through residual connections. For fair comparison, we conduct all experiments under a unified codebase that includes: the public source codes of C-HMCNN~\cite{giunchiglia2020coherent}\footnote{\href{https://github.com/EGiunchiglia/C-HMCNN}{https://github.com/EGiunchiglia/C-HMCNN}}, \revised{Soft Labels}~\cite{bertinetto2020making}\footnote{\href{github.com/fiveai/making-better-mistakes}{github.com/fiveai/making-better-mistakes}}, FGoN~\cite{chang2021your}\footnote{\href{https://github.com/PRIS-CV/Fine-Grained-or-Not}{https://github.com/PRIS-CV/Fine-Grained-or-Not}} and HRN~\cite{chen2022label}\footnote{\href{https://github.com/MonsterZhZh/HRN}{https://github.com/MonsterZhZh/HRN}}, and the reproduced codes of HD-CNN~\cite{yan2015hd} and HMCN~\cite{wehrmann2018hierarchical} since no source code is available for the original papers. When adapting these methods to semi-supervised settings, we neglect the loss over the leaf class if a sample has been relabeled to its parent class.

In \cref{tab:sota_cub}, \ref{tab:sota_air} and \ref{tab:sota_car}, we report ACC of each hierarchical level and mACC results on test sets of CUB-200-2011, Aircraft, and Stanford Cars, respectively, obtained by all competing methods. For \textbf{supervised hierarchical classification} setups (i.e., $\mathcal R=0\%$), we can observe that our LHT model consistently achieves superior or comparable performance in terms of ACC at each hierarchy, and it consistently outperforms all compared methods over mACC on all three datasets. For \textbf{semi-supervised hierarchical classification}, we select four setups (i.e., $\mathcal R=30\%, 50\%, 70\%, 90\%$) to imitate the lack of domain knowledge. Note that the dataset with $\mathcal R=30\%$ means that we randomly select 30\% samples from the training set and mask out their labels at the finest-level hierarchy during training. It can be observed that: 1) Our LHT achieves the best or second best ACC and mACC performance across all the hierarchies on all the three datasets. Particularly, in cases where limited fine-grained label information ($\mathcal R=90\%$) is available, our LHT surpasses all compared methods at all hierarchies by an evident margin, resulting in absolute mACC improvements of 5.0\%, 2.6\%, and 13.6\% compared to the second best results on all three datasets, respectively. 2) The performance of all compared methods degrade with an increasing number of relabeled samples ($\mathcal R=0\%\rightarrow90\%$), but our LHT model consistently outperforms current state-of-the-art methods. This indicates that LHT can better utilize hierarchical label information to learn a more discriminative classifier. 3) To elucidate the origins of performance improvements, we visually represent the confusion matrices obtained from the test set of Cub-200-2011 dataset, employing a relabeled ratio of $\mathcal R=90\%$. It should be noted that the diagonal vector of a confusion matrix signifies per-class recall. As shown in Fig. \ref{fig:confusion}, our LHT model exhibits a notably superior improvement compared to the baseline ResNet-50 and the current state-of-the-art method FGoN \cite{chang2021your}. Specifically, LHT achieves a mean-class-recall (defined as the arithmetic mean of recalls across all 200 classes) of 70.0\%, surpassing FGoN by 7.8\%. These results substantiate the superiority of our LHT in effectively capturing hierarchical knowledge during the training process.

\subsubsection{Ablation Study}
\label{sec:ablation}
In order to gain a deeper understanding of the effectiveness underlying the proposed LHT model, We perform extensive ablation study to evaluate and understand the contribution of each critical component in the proposed LHT model. Unless mentioned otherwise, the experiments in this subsection are all performed on Cub-200-2011 dataset with relabeled ratio $\mathcal R=90\%$.

\begin{table}
\renewcommand\arraystretch{1.2}
  \centering
  \caption{Ablation study on each key component of our LHT framework. The performance (ACC) is evaluated at the finest-level (Species) hierarchy by gradually adding Hierarchical Cross-entropy Loss (HCL), Transition Networks, and Confusion Loss into the backbone ResNet-50.}
  \label{tab:ablation}
  \resizebox{1\linewidth}{!}{
  \begin{tabular}{c|c}
    \toprule
    Model & ACC \\
    \midrule
    ResNet-50 & 55.9 \\
    ResNet-50 + HCL & 58.3 \\
    ResNet-50 + HCL + Transition Networks & 66.7 \\
    LHT (Ours) & 68.8 \\
    \hline
    \revised{LHT\_simplified (w/ Trainable Transition Matrices)} & \revised{58.1} \\
    \bottomrule
  \end{tabular}
  }
\end{table}

First, we investigate the impact of label hierarchy to demonstrate that hierarchical label information is beneficial for performance enhancement, but it requires effective learning mechanisms to exploit its underlying semantics. To this end, we append a branch for each hierarchy level on ResNet-50 and use the hierarchical cross-entropy loss (denoted as ResNet-50 + HCL) to fit the labels at each hierarchy, and the results are listed in \cref{tab:ablation}. It can be observed that the performance only exhibits a marginal improvement of about 2.4\%, indicating a limited boost in comparison to the advancements achieved by our proposed LHT model.

\begin{figure}[t]
\centering
\begin{minipage}[b]{0.47\linewidth}
    \centering
    \includegraphics[width=1.0\linewidth]{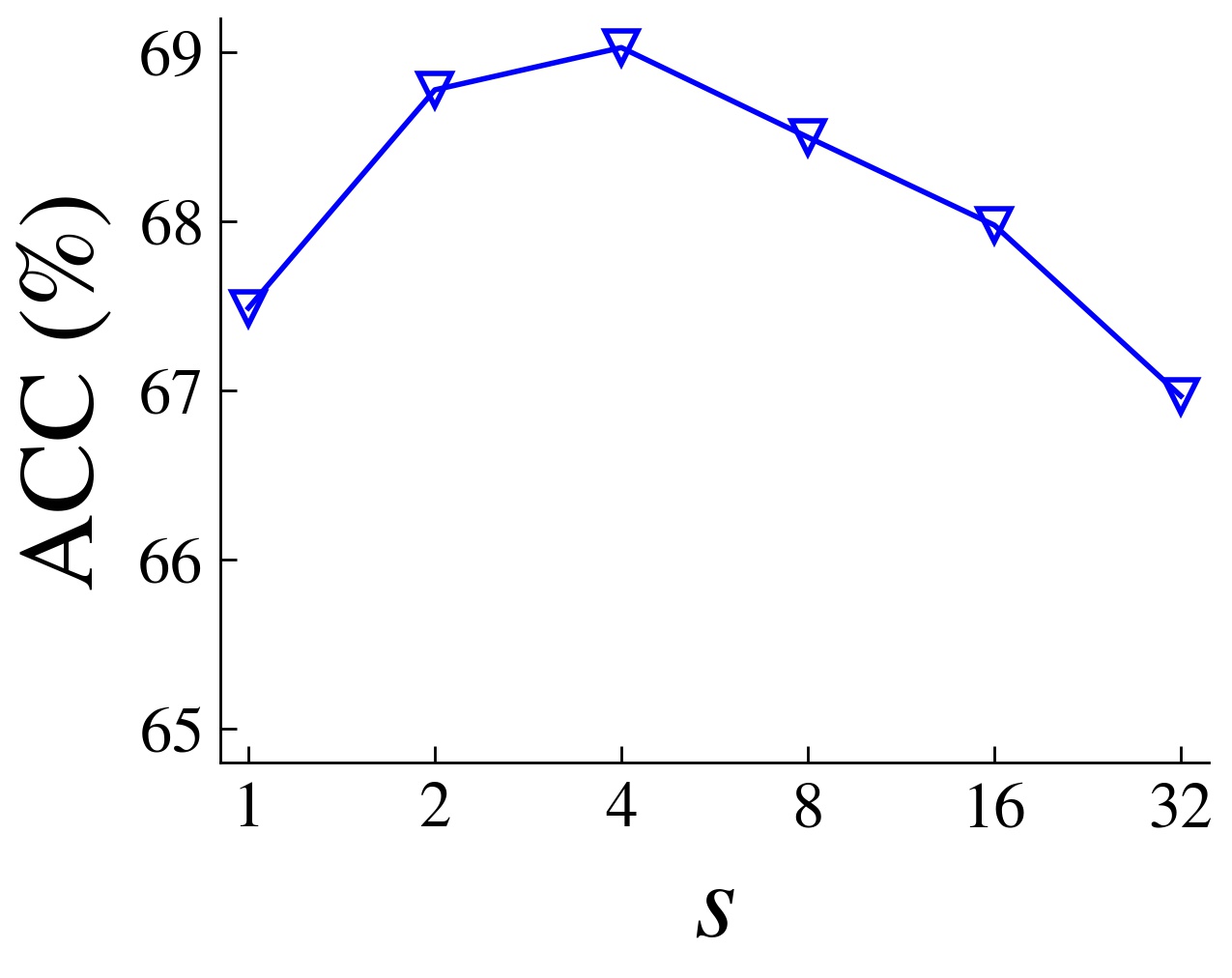}
    \caption*{(a)}
\end{minipage}
\hfill
\begin{minipage}[b]{0.47\linewidth}
    \centering
    \includegraphics[width=1.0\linewidth]{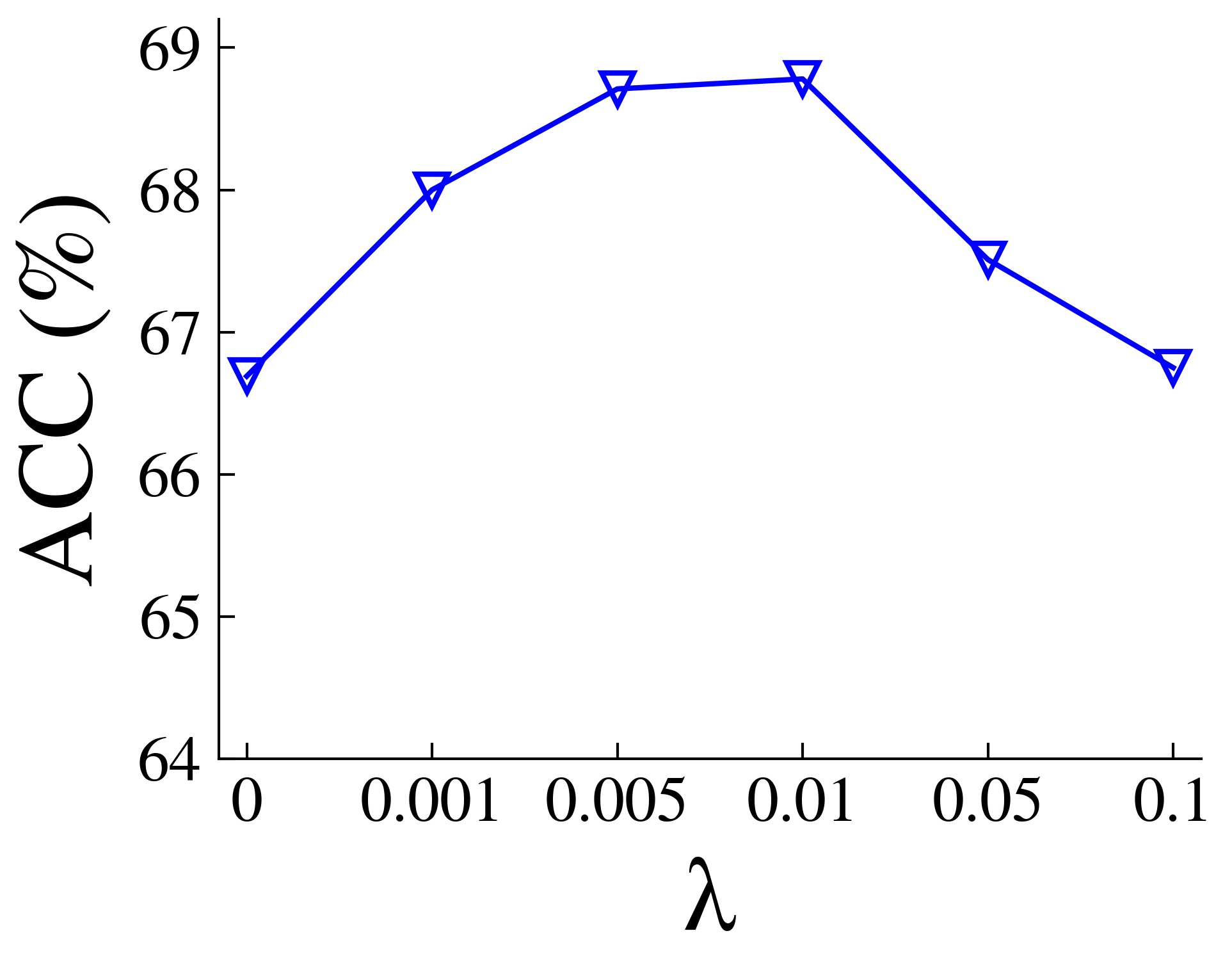}
    \caption*{(b)}
\end{minipage}
\caption{Ablation study on two hyper-parameters: (a) $s$ in the transition network (\cref{eq:low_rank_approx}) and (b) $\lambda$ in the training loss (\cref{eq:loss}).}
\label{fig:ablation}
\end{figure}


\vspace{-1mm}
\begin{table*}[t]
  \renewcommand\arraystretch{1.1}
  \centering
  \caption{Performance comparison across different backbones on CUB-200-2011. We report results on three representative CNN and ViT architectures: ResNet-50, ConvNeXt-Base, and Swin-Base. We omit the ``Base" suffix in the table.}
  \label{tab:cross_backbones}
  \resizebox{1\linewidth}{!}{
  \begin{tabular}{c| c| ccc| ccc| ccc| ccc| ccc}
    \toprule
    \multirow{2}{*}{Method} & \multirow{2}{*}{Backbone} & \multicolumn{3}{c}{$\mathcal R=0$} & \multicolumn{3}{|c}{$\mathcal R=30$} & \multicolumn{3}{|c}{$\mathcal R=50$} & \multicolumn{3}{|c}{$\mathcal R=70$} & \multicolumn{3}{|c}{$\mathcal R=90$} \\
    \cmidrule(r){3-5}\cmidrule(r){6-8}\cmidrule(r){9-11}\cmidrule(r){12-14}\cmidrule(r){15-17}
    ~ & ~ & Order & Family & Sepcies & Order & Family & Sepcies & Order & Family & Sepcies & Order & Family & Sepcies & Order & Family & Sepcies \\
    \midrule
    Baseline & ResNet-50 & 98.5 & 95.5 & 85.9 & 98.5 & 95.3 & 83.8 & 98.4 & 94.7 & 81.9 & 98.4 & 94.7 & 77.2 & 98.5 & 94.6 & 58.3\\
    Ours & ResNet-50 & \textbf{99.0} & \textbf{96.2} & \textbf{87.0} & \textbf{98.9} & \textbf{96.2} & \textbf{86.0} & \textbf{98.9} & \textbf{96.1} & \textbf{84.8} & \textbf{98.9} & \textbf{96.0} & \textbf{82.4} & \textbf{98.8} & \textbf{95.3} & \textbf{68.8} \\   
    \midrule
    Baseline & ConvNeXt & 99.2 & 96.5 & 89.2 & 99.2 & 96.4 & 88.3 & 99.2 & 96.3 & 87.0 & 99.1 & 96.0 & 83.9 & 99.0 & 96.0 & 73.3\\
    Ours & ConvNeXt & \textbf{99.4} & \textbf{97.4} & \textbf{89.3} & \textbf{99.4} & \textbf{97.3} & \textbf{88.8} & \textbf{99.4} & \textbf{97.4} & \textbf{87.8} & \textbf{99.3} & \textbf{97.1} & \textbf{85.7} & \textbf{99.3} & \textbf{96.5} & \textbf{80.1} \\
    \midrule
    Baseline & Swin & 99.2 & 96.7 & 89.2 & 99.1 & 96.6 & 88.3 & 99.1 & 96.6 & 86.5 & 99.1 & 96.4 & 83.4 & 99.0 & 95.9 & 72.4\\
    Ours & Swin & \textbf{99.5} & \textbf{97.4} & \textbf{89.5} & \textbf{99.5} & \textbf{97.2} & \textbf{88.7} & \textbf{99.5} & \textbf{97.2} & \textbf{87.5} & \textbf{99.5} & \textbf{97.1} & \textbf{85.5} & \textbf{99.3} & \textbf{96.6} & \textbf{78.9} \\   
    \bottomrule
  \end{tabular}
  }
\end{table*}

\begin{table*}[t]
  \centering
  \caption{Analysis for more sparse annotation settings on CUB-200-2011 dataset. $R^2$ denotes that a fraction of $R\%$ of the training samples lack labels at the two finest-level (Species and Family) hierarchies.}
  \label{tab:relabel}
  \resizebox{1\linewidth}{!}{
  \begin{tabular}{c |ccc| ccc| ccc| ccc}
    \toprule
    \multirow{2}{*}{Method} & \multicolumn{3}{c}{$\mathcal R=30$} & \multicolumn{3}{|c}{$\mathcal R=50$} & \multicolumn{3}{|c}{$\mathcal R=70$} & \multicolumn{3}{|c}{$\mathcal R=90$} \\
    \cmidrule(r){2-4}\cmidrule(r){5-7}\cmidrule(r){8-10}\cmidrule(r){11-13}
    ~ & Order & Family & Sepcies & Order & Family & Sepcies & Order & Family & Sepcies & Order & Family & Species \\
    \midrule
    Baseline & 98.5 & 94.5 & 83.8 & 98.4 & 93.4 & 81.4 & 98.4 & 91.1 & 75.2 & 97.9 & 82.4 & 51.0 \\
    Ours & \textbf{98.9} & \textbf{95.9} & \textbf{85.3} & \textbf{98.9} & \textbf{95.6} & \textbf{84.3} & \textbf{98.8} & \textbf{94.8} & \textbf{81.1} & \textbf{98.3} & \textbf{85.6} & \textbf{60.4} \\
    \bottomrule
  \end{tabular}
}
\end{table*}

Next, we validate the superiority of the proposed transition network, which aims to effectively encodes visual correlations embedded in class hierarchies, as delineated in Sec. \ref{sec:architecture}. To substantiate this, on one hand, we quantify the performance of the model by integrating our proposed Transition Networks into the baseline model (ResNet-50 + HCL). As shown in \cref{tab:ablation}, the model achieves an accuracy of 66.7\%, which yields an absolute gain of 8.4\% over the baseline model. This substantiates the efficacy of our proposed transition network. We further investigate the hyper-parameter $s$ introduced by the Transition Networks in Fig. \ref{fig:ablation}(a). We can see that the performance remains relatively stable as the value of $s$ varies. As such, we fix it as 2 across all our experiments for a better trade-off between computation efficiency and model performance. On the other hand, if we directly treat the transition matrices as trainable parameters (LHT\_simplified) instead of learning them through the Transition Networks, model performance degrades from 68.8\% to 58.1\%, further validating the effectiveness of the Transition Networks.

Lastly, we conduct ablation analysis on the proposed confusion loss. As shown in \cref{tab:ablation}, when the confusion loss is integrated into the model (ResNet-50 + HCL + Transition Networks), the resulting LHT model achieves a further 2.1\% performance improvement. This verifies the effectiveness of the proposed confusion loss on encouraging the transition network to capture the semantic correlation within class hierarchies. In fact, the hyper-parameter $\lambda$ in \cref{eq:loss} plays an important role in trading off between the hierarchical cross-entropy loss and the confusion loss during training, and we thus conduct a detailed ablation analysis on $\lambda$ in Fig. \ref{fig:ablation}(b). It can be observed that the initial performance gradually improves as $\lambda$ increase from 0 to 0.01, which suggests that a smaller $\lambda$ facilitates the transfer network to output meaningful transfer matrices; however, a much larger $\lambda$ excessively forces the output transfer matrices to be more smooth, which may amplify biases introduced by inaccurate predictions, especially in semi-supervised scenarios.

\subsubsection{Discussion and Analysis}
We provide a comprehensive discussion from multiple perspectives to enrich the understanding of our approach by addressing the following questions.

\medskip
\noindent\textbf{How does LHT generalize across various architectures?} In addition to ResNet-50, we further evaluate our method on ConvNeXt-Base and Swin-Base, which are representative modern CNN and ViT architectures. Compared to ResNet-50, these models have substantially larger parameter capacities, extending the evaluation to a much broader capacity range. We keep all other experimental settings consistent with those used for ResNet-50. As shown in Tab. 5, our proposed LHT consistently outperforms the baseline HD-CNN \cite{yan2015hd} that employs independent classification heads for each hierarchy under varying missing-label ratios. These results demonstrate that our approach generalizes well to both advanced CNNs and ViT-based architectures, confirming its robustness across diverse architecture designs.

\medskip
\noindent\textbf{How does LHT perform under sparser annotation settings?} As aforementioned, our LHT presents an effective learning mechanism to exploit hierarchical label information, especially in cases where a significant portion of labels are missing at the finest-level hierarchy. Herein, we further validate the efficacy of our method in a more challenging semi-supervised scenario, where we randomly select a fraction of samples and mask out their labels at the two finest-level hierarchies (i.e., Species and Family) on Cub-200-2011 dataset. As shown in \cref{tab:relabel}, Our proposed LHT model consistently improves the baseline HD-CNN \cite{yan2015hd} across different amounts of relabeled data. For example, in an extremely sparse annotation scenario ($R^2=90\%$), LHT significantly improves the baseline model by around 9.4\% ACC at the finest-level hierarchy and 4.3\% mACC across all hierarchies. This demonstrates that our LHT achieves higher performance with less label information, thus confirming its superiority in effectively utilizing class hierarchies.

\begin{figure}[t]
\centering
\begin{minipage}[b]{1.0\linewidth}
    \centering
    \includegraphics[width=1.0\linewidth]{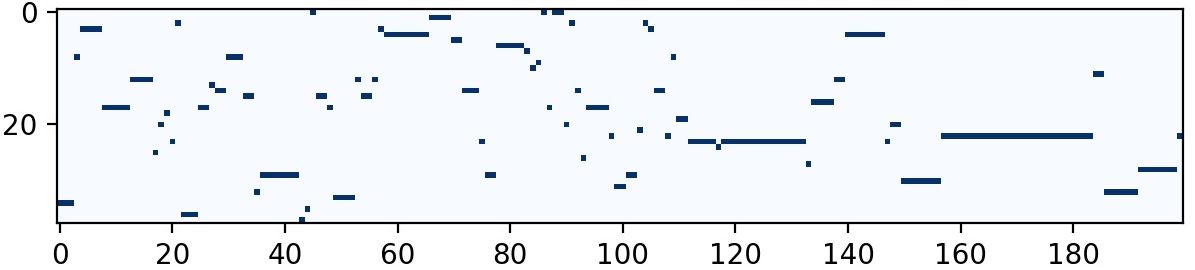}
\end{minipage}
\hfill\vspace{0.5mm}
\begin{minipage}[b]{1.0\linewidth}
    \centering
    \includegraphics[width=1.0\linewidth]{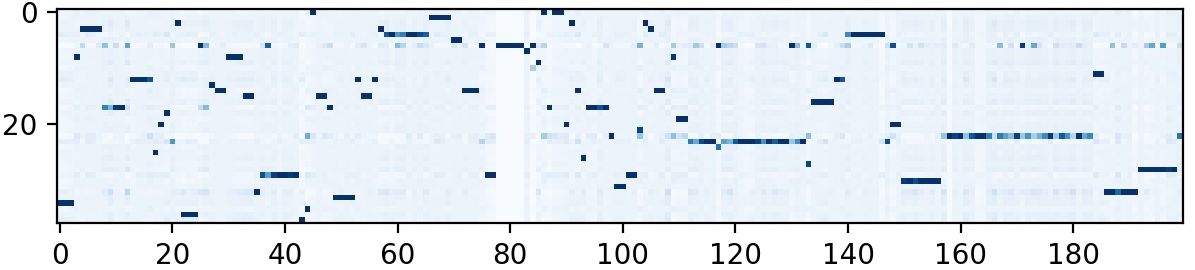}
\end{minipage}
\caption{Visualization of transition matrices (\textit{Species} $\to$ \textit{Family}) on CUB-200-2011: the top shows the naive transition matrix $S^{1}$ inferred from hierarchical labels, and the bottom shows the learned transition matrix $T^{1}$.}
\label{fig:transition}
\end{figure}

\medskip
\noindent\textbf{Does the learned transition matrix encode semantic correlations?} To answer this question, we visualize the transition matrices output by the Transition Network on Cub-200-2011 in \cref{fig:transition}. Specifically, we show the learned transition matrix associated with the two finest-level hierarchies, denoted as $\mathbf T^1 (species\rightarrow family)$, and also visualize the corresponding naive transition matrices $\mathbf S^1 (species\rightarrow family)$ directly inferred from hierarchical labels (as mentioned in \textbf{Remark 1}) for better reference. We can see that the naive transition matrix can only encode the deterministic state transition from the finer- to the coarser-level hierarchies since only one element is 1 in each column, whereas the ones learned from our proposed LHT model are indeed conditional label distributions considering the uncertainty from image context and hierarchy prior. This is mainly because, on the one hand, the learned matrices largely reflect the transition states in the naive transfer matrix, while on the other hand, LHT produces distinct transition matrices for different input samples. We also visualize the evolution of the weight $\gamma$ during training in Appendix E to clarify the impact of the learned transition matrices.

\begin{figure}[t]
    \centering
    \includegraphics[width=0.8\linewidth]{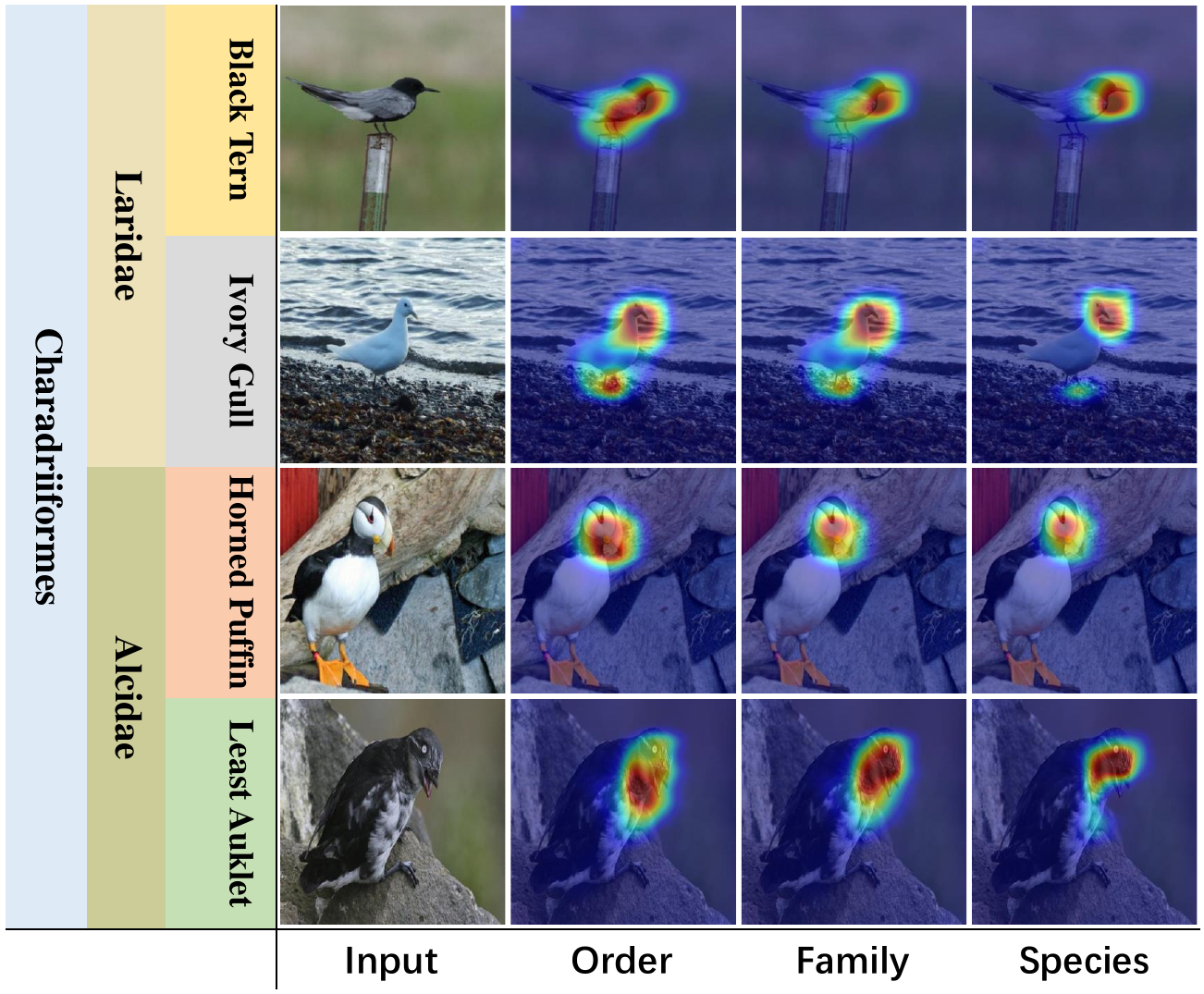}
    \caption{Visualization of the attention maps returned by Grad-Cam \cite{selvaraju2017grad} of the three hierarchical branches of the proposed LHT model, with each one corresponding to a specific hierarchy. The leftmost column is the original image, and the three columns on the right correspond to the attention maps of Order, Family and Species hierarchy, respectively.}
    \label{fig:hierarchy_cam}
\end{figure}

\medskip
\noindent\textbf{Does the learned attention map encode semantic correlations?} In Fig. \ref{fig:hierarchy_cam}, we further employ Grad-Cam \cite{selvaraju2017grad} to visualize the attention regions of each hierarchical branch by propagating their respective gradients back to feature maps for four typical examples. These objects belong to the same order (i.e., \emph{``Charadriiformes"}), two families (i.e., \emph{``Laridae"} and \emph{``Alcidae"}), and four species (i.e., \emph{``Black Tern", ``Ivory Gull", ``Least Auklet"} and \emph{``Horned Puffin"}). It can be observed that 1) our proposed LHT model exhibits a preference for activating more global attention regions in the coarser-level hierarchy compared to the finer-level hierarchy, and the visualization results demonstrate notable difference between the attention patterns of the finer-level categories and that of the coarser-level categories; 2) our LHT model effectively directs attention towards discriminative regions when classifying examples from different species within the same family. For example, when distinguishing between the species \emph{``Least Auklet"} and \emph{``Horned Puffin"}, both belonging to the family \emph{``Alcidae"}, the model focuses intensely on the head, particularly the eye and beak regions, which provide essential discriminative information. This further validates the capability of our LHT model in capturing semantic correlations across different hierarchies to some extent.

\subsection{Low-shot Learning with Class Hierarchy}
Large-scale low-shot learning (LSL) involves a large number of source classes with abundant labeled data and target classes with only a few labeled examples. The goal is to recognize unlabeled target samples, which is challenging due to their limited supervision and the difficulty of directly learning discriminative classifiers.
\subsubsection{Problem Formulation}
Formally, let $S_{source}$ denote the set of source classes and $S_{target}$ denote the set of target classes. The two class sets satisfy the following properties: 1) $S_{source}$ and $S_{target}$ are disjoint, i.e., $S_{source}\cap S_{target}=\emptyset$; 2) the semantic correlation of $S_{source}$ and $S_{target}$ are explicitly encoded into a tree-structured class hierarchy, where the leaf nodes cover all source and target classes and the parent nodes group semantically related leaf classes. We are given a large-scale source set $\mathcal D_{source}$ from source classes $S_{source}$, a few-shot support set $\mathcal D_{target}$ from target classes $S_{target}$, and a test set $\mathcal D_{test}$ from target classes $S_{target}$. In this setting, $\mathcal D_{source}$ contains sufficient labeled samples for each source class, while $\mathcal D_{target}$ provides only a few (e.g., 5) labeled samples per target class. The objective is to predict labels for $\mathcal D_{test}$.

Following \cite{li2019large}, we address the challenge of LSL by introducing hierarchical classification. The key idea is that, although source and target classes are disjoint, they often share higher-level categories in a tree-structured hierarchy. For example, certain source and target classes may share the broader category of ``vehicles", with the former covering various vehicle types and the latter including specific categories such as cars. This shared structure provides a natural bridge between the source and target classes. By performing a hierarchical classification on the source data, the network learns transferable features that capture such semantic relationships. These representations serve as informative priors from semantically related source classes and are particularly beneficial for recognizing target classes.

Specifically, the method follows a two-stage process: \romannumeral1) Training stage, in which only source class samples are used to train the hierarchical network. Since the source classes and target classes share certain common super-classes, the network facilitates the learning of transferable features for LSL on the target classes. \romannumeral2) Test stage, in which visual features of the few-shot support samples and the test samples (both from target classes) are extracted through the learned hierarchical classification network. The test samples are then recognized by a simple nearest neighbor search using the visual features of the support samples as references. Specifically, when adapting the proposed LHT method to large-scale LSL, we train the hierarchical classification network with the source data in the training stage and employ the features from global average pooling as input for the nearest-neighbor classifier in the test stage.

\subsubsection{Experimental Setups}
A benchmark derived from ILSVRC2012/2010 \cite{deng2009imagenet} is used for performance evaluation \cite{li2019large}. This dataset consists of three parts: a support set with 1000 source classes from ILSVRC2012, a few-shot support set with 360 target classes from ILSVRC2010 (not included in ILSVRC2012), and a test set containing target class samples excluding those in the support set. Following \cite{li2019large}, we construct a class hierarchy with three levels according to: 1) we first cluster the word vectors\footnote{The word vectors are obtained by a skip-gram text model \cite{mikolov2013distributed} in \cite{li2019large} and available at \href{https://github.com/tiangeluo/fsl-hierarchy}{https://github.com/tiangeluo/fsl-hierarchy}.} of source/target class names using k-means to create a coarser-level hierarchy, and the super-class word vectors are obtained by averaging the word vectors of their child classes; 2) we then acquire upper-layer nodes by clustering the word vectors of the lower-layer nodes. These three hierarchies have 1000, 100, and 10 classes, respectively. We refer to \cite{li2019large} and employ Momentum SGD with a momentum of 0.9, a weight decay of $5\times10^{-4}$, a batch-size of 128, and an initial learning rate of 0.001 for backbone layers and 0.01 for other layers to train the model for 5 epochs. The top-5 accuracy on $\mathcal D_{test}$ is adopted as the evaluation metric.

\subsubsection{Experimental Results}

\begin{table}
  \renewcommand\arraystretch{1.1}
  \centering
  \caption{Results for large-scale LSL on ILSVRC2012/2010 dataset. The best results are highlighted in \textbf{bold}.}
  \label{tab:few_shot}
  \setlength{\tabcolsep}{2.0mm}{
  \begin{tabular}{l| c cc cc}
    \toprule
    \multirow{2}{*}{Methods} & \multicolumn{5}{c}{$K$-shot} \\
    \cmidrule(r){2-6}
    ~ & $K=1$ & $K=2$ & $K=3$ & $K=4$ & $K=5$ \\
    \midrule
    NN & 34.2 & 43.6 & 48.7 & 52.3 & 54.0 \\
    SGM \cite{hariharan2017low} & 31.6 & 42.5 & 49.0 & 53.5 & 56.8 \\
    PPA \cite{qiao2018few} & 33.0 & 43.1 & 48.5 & 52.5 & 55.4 \\
    LSD \cite{douze2018low} & 33.2 & 44.7 & 50.2 & 53.4 & 57.6 \\
    KT \cite{li2019large} & 39.0 & 48.9 & 54.9 & 58.7 & 60.5 \\
    Ours & \textbf{41.9} & \textbf{52.2} & \textbf{58.6} & \textbf{61.6} & \textbf{63.7} \\
    \bottomrule
  \end{tabular}
  }
\end{table}

We compare our model with five large-scale LSL models: (\romannumeral1) NN, a baseline model employing ResNet50 pre-trained on $\mathcal D_{source}$ as the nearest neighbor (NN) classifier; (\romannumeral2) SGM \cite{hariharan2017low}, a LSL model using the squared gradient magnitude (SGM) loss [7]; (\romannumeral3) PPA \cite{qiao2018few}, the parameter prediction with activation (PPA) model; (\romannumeral4) LSD \cite{douze2018low}, a LSL model with large-scale diffusion (LSD); (\romannumeral5) KT \cite{li2019large}, the state-of-the-art LSL model via knowledge transfer (KT) with class hierarchy. The results in \cref{tab:few_shot} show that our proposed LHT method achieves the best result among all the comparison methods, and the performance gain is significant compared to KT. While KT also introduces hierarchical classification by feeding fine-grained predictions into coarser-level classifiers, it does not explicitly model statistical dependencies across hierarchies. Instead, our LHT makes these dependencies explicit by formulating inter-level correlations as conditional probabilities, which facilitates the capture of hierarchical relationships during training. This design yields more transferable features and explains the consistent improvements compared to KT.

\section{Experiments on Medical Images}
\label{sec:extension}
In this section, we empirically shed light on the potential of the proposed LHT applied in the domain of computer-aided diagnosis. Concretely, we reformulate the lesion diagnosis problem into a hierarchical classification task, as we can readily obtain a hierarchical categorization of lesions based on pathological priors \cite{esteva2017dermatologist,barata2019deep}.


\begin{table}[t]
\renewcommand\arraystretch{1.8}
\setlength{\tabcolsep}{6pt}
\centering
\caption{Summary of the ISIC2018 dataset. The skin lesion categories are organized into three hierarchical levels.}
\label{tab:skin}
\resizebox{1\linewidth}{!}{
\begin{tabular}{l| c c c c c c c}
\toprule
\multirow{3}{*}{\makecell{Class\\Hierarchy}} &
\multicolumn{2}{c}{Melanocytic} &
\multicolumn{5}{c}{Non-melanocytic} \\
\cmidrule(lr){2-3}\cmidrule(lr){4-8}
& Benign & Malignant & \multicolumn{3}{c}{Benign} & \multicolumn{2}{c}{Malignant} \\
\cmidrule(lr){2-2}\cmidrule(lr){3-3}\cmidrule(lr){4-6}\cmidrule(lr){7-8}
& NV & MEL & BKL & VASC & DF & BCC & AKIEC \\
\hline
Number & 6705 & 1113 & 1098 & 143 & 115 & 514 & 327 \\
\bottomrule
\end{tabular}
}
\end{table}

\begin{table*}
  \renewcommand\arraystretch{1.1}
  \centering
  \caption{Per-class recall ($\%$) on the test set of ISIC2018 dataset. $R$ and $R^2$ represent the random selection of $R\%$ training samples for label masking at the finest-level and two finest-level hierarchies, respectively.}
  \label{tab:skin_1st}
  \setlength{\tabcolsep}{3.5mm}{
  \begin{tabular}{l cc cccccccc c}
    \toprule
    \multirow{2}{*}{Method} & \multicolumn{2}{c}{Training Data} & \multicolumn{7}{c}{Lesion Category} & \multirow{2}{*}{Avg.} \\
    \cmidrule(r){2-3}\cmidrule(r){4-10}
    ~ & Labeled & Patially-labeled & NV & MEL & BKL & BCC & AKIEC & VASC & DF & ~ \\
    \midrule
    ResNet-50 & $20\%$ & 0 & 94.1 & 42.4 & 60.1 & 68.7 & 68.7 & 77.9 & 58.2 & 67.2 \\
    \hline
    LHT (Ours) & $20\%$ & $\mathcal R=80\%$ & 97.1 & 63.3 & 80.4 & 87.7 & 78.8 & 89.7 & 80.0 & 82.4\\
    LHT (Ours) & $20\%$ & $R^2=80\%$ & 96.4 & 42.7 & 74.1 & 81.0 & 77.6 & 86.2 & 72.9 & 75.8 \\
    \bottomrule
  \end{tabular}
  }
\end{table*}

\begin{table*}
  \renewcommand\arraystretch{1.1}
  \centering
  \caption{Comparison results over Precision, Recall, Accuracy, F1-Score and AUC on ISIC2018 dataset. $R$ and $R^2$ represent the random selection of $R\%$ training samples for label masking at the finest-level and two finest-level hierarchies, respectively. The best and second best results are highlighted in \textbf{bold} and \underline{underline}, respectively.}
  \label{tab:skin_2nd}
  \setlength{\tabcolsep}{2.5mm}{
  \begin{tabular}{l ccccc ccccc}
    \toprule
    \multirow{2}{*}{Method} & \multicolumn{5}{c}{ISIC2018 ($\mathcal R=80\%$)} & \multicolumn{5}{c}{ISIC2018 ($R^2=80\%$)} \\
    \cmidrule(r){2-6} \cmidrule(r){7-11}
    ~ & Precision & Recall & Accuracy & F1-Score & AUC & Precision & Recall & Accuracy & F1-Score & AUC \\
    \midrule
    HD-CNN \cite{yan2015hd} & 79.8 & \underline{79.9} & \underline{89.8} & \underline{79.4} & 97.3      & 73.4 & 73.0 & 84.9 & 72.3 & 95.1 \\
    HMCN \cite{wehrmann2018hierarchical} & 78.5 & 72.2 & 86.6 & 74.3 & 94.5    & 66.6 & 48.4 & 78.0 & 53.7 & 85.9 \\
    C-HMCNN \cite{giunchiglia2020coherent} & \underline{80.2} & 78.2 & 89.4 & 78.8 & \underline{97.5}    & \underline{73.6} & 72.4 & \underline{85.4} & 72.1 & \underline{95.8} \\
    FGoN \cite{chang2021your} & 78.8 & 78.2 & 88.5 & 77.8 & 96.9     & 72.9 & \underline{73.7} & 84.9 & \underline{72.7} & 95.1 \\
    HRN \cite{chen2022label} & 77.7 & 77.6 & 88.6 & 77.1 & 96.4    & 72.8 & 71.5 & 83.9 & 71.8 & 94.3 \\
    LHT (Ours) & \textbf{81.3} & \textbf{82.4} & \textbf{90.1} & \textbf{81.3} & \textbf{98.2}   & \textbf{76.4} & \textbf{75.8} & \textbf{86.1} & \textbf{75.3} & \textbf{96.7} \\
    \bottomrule
  \end{tabular}
  }
\end{table*}

\subsection{Experimental Setups}
\textbf{Dataset.} The proposed model is evaluated on the publicly available ISIC2018 dataset \cite{tschandl2018ham10000} from the ISIC2018 challenge on skin lesion diagnosis\footnote{\href{https://challenge2018.isic-archive.com}{https://challenge2018.isic-archive.com}}. It is a challenging dataset due to the class-imbalanced issue. The dataset contains totally 10015 dermoscopic images for seven skin diseases, namely Melanocytic Nevi (NV), Melanoma (MEL), Benign Keratosis (BKL), Basal Cell Carcinoma (BCC), Actinic Keratosis / Bowens disease (AKIEC), Vascular Lesion (VASC), and Dermatofibroma (DF). We organize these seven skin lesions into three hierarchies according to their origin (melanocytic or non-melanocytic), degree of malignancy (benign or malignant), and finally their diagnosis (e.g., melanoma, basal cell carcinoma, vascular lesion, etc.) following \cite{barata2019deep, argenziano2000interactive}. In \cref{tab:skin}, we list the statistics and class hierarchies of the dataset. Since the ground truth of official validation and testing set was not released, we randomly split the entire training set to 70\% for training, 10\% for validation and 20\% for testing. To verify the proposed method can notably improve the performance of skin lesion diagnosis and concurrently reduce the annotation cost of training samples, we conduct experiments in a semi-supervised learning regime, i.e., randomly selecting 80\% of training samples and masking out their labels at the finest-level hierarchy ($\mathcal R=80\%$) or the two finest-level hierarchies ($R^2=80\%$).

\medskip
\noindent\textbf{Evaluation Metrics.} To quantitatively evaluate the proposed method, we adopt five metrics, including Precision, Recall, F1-score, AUC\footnote{Each metric is calculated as the average value across all classes.} and Accuracy for evaluation. We report the averaged results from five
independent runs with different random seeds.

\medskip
\noindent\textbf{Implementation details.} We employ ResNet-50 pre-trained on ImageNet \cite{deng2009imagenet} as our network backbone, and train the models with Momentum SGD with a momentum of 0.9, a weight decay of $5\times10^{-4}$, a batch-size of 8, and an initial learning rate of 0.0002 for backbone layers and 0.002 for other layers. The training epoch is set as 200 and the learning rate is decayed by the cosine annealing schedule. The hyper-parameter $\lambda$ in Eq. \ref{eq:loss} is set as 0.01. In all our experiments, we resize input images to $448 \times 448$ (originally around 600 × 450 pixels) and simply apply random horizontal flipping and random cropping (random cropping for training and center cropping for testing) for data augmentation.

\subsection{Experimental Results}
To understand the effectiveness of the proposed LHT applied to skin lesion diagnosis, we herein answer the following research questions under both $\mathcal R=80\%$ and $R^2=80\%$ settings: 1) Does LHT contribute to the diagnosis of skin lesions by using class hierarchy? 2) How does LHT perform compared with prior hierarchical classification methods when used in skin lesion diagnosis?

To answer the first question, we report the per-class recall results for all the seven classes and compare it with a vanilla model (ResNet-50 without using hierarchical label information to train the model) in \cref{tab:skin_1st}. It can be observed that: 1) Our proposed LHT method consistently improves the recall of each lesion category and improve the average recall by $67.2\%\rightarrow 82.4\%$ under $\mathcal R=80\%$ and $67.2\%\rightarrow 75.8\%$ under $R^2=80\%$, respectively. This demonstrates that our approach effectively leverages semantic correlations within the label hierarchy to enhance the performance of skin lesion diagnosis. 2) In the setting of $R^2=80\%$, it only requires coarse-level annotations (just identifying whether the lesion is melanocytic or non-melanocytic) for 80\% of training data, and our LHT model results in an absolute performance gain of up to 8.6\%. This highlights the significant potential of our method to reduce the annotation cost for computer-aided diagnosis systems.

In \cref{tab:skin_2nd}, we further compare the proposed LHT with state-of-the-art methods for skin lesion diagnosis on ISIC2018 dataset. The comparison methods include HD-CNN~\cite{yan2015hd}, HMCN~\cite{wehrmann2018hierarchical}, C-HMCNN~\cite{giunchiglia2020coherent}, FGoN~\cite{chang2021your}, and HRN~\cite{chen2022label}, which are described in detail in \cref{sec:comparision}. The results show that our LHT model consistently outperforms all comparison methods over all five evaluation metrics. For example, we surpass the second best result by $79.9\%\rightarrow 82.4\%$ under $\mathcal R=80\%$ and $73.7\%\rightarrow 75.8\%$ under $R^2=80\%$ on recall, highlighting that the proposed LHT method indeed helps to exploit the coarsely annotated hierarchical labels for skin lesion diagnosis. This reveals that our method can effectively leverage the training data annotated at different hierarchical levels to train a computer-aided diagnosis system. As such, the annotators may only need to provide partial labels with the highest confidence based on their own expertise, rather than annotating every sample into the finest-level label, which largely alleviates the conformation bias arising from noisy labels during training.

\section{Conclusion and Discussion}
\label{sec:conclusion}
In this work we have proposed a probabilistic approach to address the hierarchical classification problem. Our key insight is to directly encode the correlation between two adjacent class hierarchies as conditional probabilities, such that the category labels across different hierarchies can be recursively predicted in a fine-to-coarse order. Beyond that, we have designed a confusion loss to further encourage the deep classifiers to learn the hierarchical correlation. We demonstrate the superiority of our method in hierarchical classification on a series of benchmark datasets and further shed light on the potential of the proposed method in computer-aided diagnosis.

Despite these promising results, our method has certain limitations. First, it relies on predefined hierarchical structures, which may not always be available in real-world applications. Second, the current framework focuses primarily on single-label classification, and its scalability to multi-label settings or to more general graph-structured hierarchies (e.g., cases where certain nodes have multiple parent categories) remains to be fully investigated. These limitations point to several promising directions for future research, including incorporating automatically discovered or partially observed hierarchies and extending the framework to multi-label and graph-structured scenarios.


%

\appendices
\section{Theoretical Proofs in Section 3}
\subsection{Derivation of Eq. (1) Equivalent to Solving a Negative Log-Likelihood}
\label{app:lle}
In this section, we provide a detailed derivation that establishes the equivalence between the hierarchical cross-entropy loss $\mathcal L_{{\rm CE}}=\sum_{n=1}^N\sum_{k=1}^K H(\mathbf p^k_n, \mathbf y^k_n)$ presented in Eq. (1) of the main text and the negative log-likelihood $p(y^1,\cdots,y^K|\mathcal I)=\prod_{k=1}^K p(y^k|\mathcal I)$.

\medskip
\noindent\textit{Derivation.} We first denote $\mathbf y^k$ as the one-hot vector corresponding to $y^k$, $\mathbf p^k$ as the associated prediction for $y^k$, and $\mathbf v[i]$ as the $i$-th entry of the vector $\mathbf v$. We then have

\begin{equation}
\begin{aligned}
-\log p(y^1,\cdots,y^K|\mathbf x) = &-\log\prod_{k=1}^K p(y^k|\mathbf x) \\
=&-\sum_{k=1}^K \log p(y^k|\mathbf x) \\
=&-\sum_{k=1}^K \log \prod_{i=1}^{C_k}(\mathbf p^k[i])^{\mathbf y^k[i]} \\
=&-\sum_{k=1}^K \sum_{i=1}^{C_k}\mathbf y^k[i]\log \mathbf p^k[i] \\
=&\sum_{k=1}^K H(\mathbf y^k, \mathbf p^k).
\end{aligned}
\end{equation}
Note that to the right of the third equal sign, it implies that the prediction at each hierarchy follows a categorical distribution.

\subsection{Proof of Proposition 1}
\addtocounter{proposition}{-2}
\begin{proposition}
In the LHT framework, the label prediction of the $k$-th ($k>1$) hierarchy can be inferred from the finest-level hierarchy as follows:
\begin{equation}
\tag{9}
\mathbf p^k={\mathbf T}^{k\leftarrow1} {\mathbf p}^{1},
\end{equation}
where the transition matrix $\mathbf{T}^{k \leftarrow 1}$ is defined as $\mathbf{T}^{k \leftarrow 1} := \mathbf{T}^k \mathbf{T}^{k-1} \cdots \mathbf{T}^2$, which is implemented by the composite function $g^k_{\theta^k} \circ g^{k-1}_{\theta^{k-1}} \circ \cdots \circ g^2_{\theta^2}$.
\end{proposition}

\begin{proof}
According to Eq. (2) in the main text, we have
\begin{equation}
    \label{eq:forward_transfer}
    \mathbf p^k=\mathbf T^k \mathbf p^{k-1}=\mathbf T^k \mathbf T^{k-1}\mathbf p^{k-2}=\cdots=(\mathbf T^k\cdots\mathbf T^2) \mathbf p^1.
\end{equation}
Let \(\mathbf T^{k\leftarrow1}=(\mathbf T^k\cdots\mathbf T^2)\), which gives us \(\mathbf p^k={\mathbf T}^{k\leftarrow1} {\mathbf p}^{1}\). To complete the proof, we need to verify that each entry of \(\mathbf T^{k\leftarrow1}\) represents a transition probability. In fact, we can demonstrate that $(t^{k\leftarrow1}_{ij}=p(Y^k=i|Y^1=j,\mathbf x)$ using the principle of mathematical induction.

The case for \(k=2\) is clearly valid, as we have $\mathbf T^{2\leftarrow 1}=\mathbf T^2$ and
\begin{equation}
t^{2\leftarrow 1}_{ij}=t^2_{ij}=p(Y^2=i|Y^1=j,\mathbf x).
\end{equation}

Assuming that the statement holds for $k-1$, we have
\begin{equation}
\mathbf T^{(k-1)\leftarrow 1}=\mathbf T^{k-1}\cdots\mathbf T^2,
\end{equation}
and 
\begin{equation}
t^{(k-1)\leftarrow1}_{ij}=p(Y^{k-1}=i|Y^1=j,\mathbf x).
\end{equation}
Next, we consider the case for $k$. According to \cref{eq:forward_transfer}, we can express this as:
\begin{equation}
    \mathbf T^{k\leftarrow 1}=\mathbf T^k \mathbf T^{(k-1)\leftarrow 1}.
\end{equation}
Its $(i,j)$-th entry $t_{ij}^{k\leftarrow1}$ can be computed as follows:
\begin{equation}
\begin{aligned}
    t_{ij}^{k\leftarrow1} & =\sum_{n=1}^{C_{k-1}} t^k_{in} t^{(k-1)\leftarrow1}_{nj} \\
    & =\sum_{n=1}^{C_{k-1}} p(Y^k=i|Y^{k-1}\!=\!n,\mathbf x) \cdot p(Y^{k-1}\!=\!n|Y^1\!=\!j,\mathbf x) \\
    & =\sum_{n=1}^{C_{k-1}} p(Y^k=i, Y^{k-1}=n|Y^1=j, \mathbf x) \\
    & = p(Y^k=i|Y^1=j, \mathbf x).
\end{aligned}
\end{equation}
This confirms that the statement holds for the case of $k$. By the principle of induction, we conclude that the statement is true for all $k \ge 2$, thereby we complete the proof.
\end{proof}

\subsection{Proof of Proposition 2}
\begin{proposition}[]
\label{thr:backward_transfer}
If the loss function in Eq.(7) are minimized by gradient based algorithms (such as SGD), then the loss term $\mathcal L^k\triangleq H(\mathbf p^k, \mathbf y^k)$ enables the propagation of true label information from coarse-level hierarchies to the finest-level one through gradient backpropagation.
Specifically, the gradient of $L^k$ with respect to the logits $\mathbf{z}$ of the finest-level hierarchy is computed as
\begin{equation*}
\tag{10} \label{eq:backward_transfer}
\begin{aligned}
    \mathcal G^k_j := \frac{\partial\mathcal L^k}{\partial \mathbf z[j]} =\bigl(1-\frac{t^{k\leftarrow{1}}_{y^k j}}{\mathbf p^k[y^k]}\bigr)\mathbf p^1[j],
\end{aligned}
\end{equation*}
where $\mathbf z[j]$ denotes the $j$-th entry of $\mathbf z$ and $k=2,\cdots,K$.
\end{proposition}

\begin{proof}
Our proof comprises two primary parts: First, we demonstrate that $\mathcal L^k \triangleq H(\mathbf p^k, \mathbf y^k)$ effectively facilitates the transfer of true label information from coarse-level hierarchies to the finest-level hierarchy through gradient backpropagation. Secondly, we derive the formulation in \cref{eq:backward_transfer}, i.e., the gradient of $\mathcal L^k$ with respect to the logits $\mathbf z$ of the finest-level hierarchy.

\medskip
\noindent\textbf{(Part I)} We begin by demonstrating that if the network parameters are updated according to Eq. (7) using SGD-based optimizers, the loss term $\mathcal L^k \triangleq H(\mathbf p^k, \mathbf y^k)$ effectively facilitates the transfer of information from coarse-level hierarchies to the finest-level hierarchy through gradient backpropagation. 

During the backpropagation phase, the we can easily obtain that
\begin{equation}
    \label{eq:pro2_1}
    \begin{aligned}
    \frac{\partial \mathcal L^k}{\partial \mathbf p^1} & = \frac{\partial H \bigl({\mathbf p}^{k}, \mathbf y^k\bigr)}{\partial \mathbf p^1} = \frac{\partial H \bigl({\mathbf p}^{k}, \mathbf y^k\bigr)}{\partial \mathbf p^k} \frac{\partial \mathbf p^k}{\partial \mathbf p^1} \\
    & = \frac{\partial H \bigl({\mathbf p}^{k}, \mathbf y^k\bigr)}{\partial \mathbf p^k} \frac{\partial \mathbf p^k}{\partial \mathbf p^{k-1}} \frac{\partial \mathbf p^{k-1}}{\partial \mathbf p^{1}} \\
    & = \cdots = \frac{\partial H \bigl({\mathbf p}^{k}, \mathbf y^k\bigr)}{\partial \mathbf p^k} \frac{\partial \mathbf p^k}{\partial \mathbf p^{k-1}} \frac{\partial \mathbf p^{k-1}}{\partial \mathbf p^{1}} \cdots \frac{\partial \mathbf p^3}{\partial \mathbf p^2} \frac{\partial \mathbf p^2}{\partial \mathbf p^1},
    \end{aligned}
\end{equation}
where the second equation holds due to the chain rule; the third equation is valid because $\mathbf p^k = \mathbf T^k \mathbf p^{k-1}$, which means $\mathbf p^k$ is a function of $\mathbf p^{k-1}$. By the chain rule, it then follows that \(\frac{\partial \mathbf p^k}{\partial \mathbf p^1} = \frac{\partial \mathbf p^k}{\partial \mathbf p^{k-1}} \frac{\partial \mathbf p^{k-1}}{\partial \mathbf p^1}\).

From \cref{eq:pro2_1}, it is evident that the gradient flow originating from hierarchy $k$ sequentially propagates through hierarchies $k-1$, $k-2$, and continues down to the finest-level hierarchy. This indicates that the loss term $\mathcal{L}^k \triangleq H(\mathbf{p}^k, \mathbf{y}^k)$ facilitates the transfer of information from coarser-level hierarchies to the finest-level hierarchy via gradient backpropagation. Thus, we have substantiated this conclusion.

\medskip
\noindent\textbf{(Part II)} In the following, we drive the gradient of $\mathcal L^k$ with respect to the logits $\mathbf z$ of the finest-level hierarchy, as outlined in \cref{eq:backward_transfer}.

We start by computing the derivative of the loss term $\mathcal L^k\triangleq H(\mathbf p^k, \mathbf y^k)$ with respect to the logits $\mathbf z$ of the finest-level hierarchy:
\begin{equation}
\begin{aligned}
    \frac{\partial \mathcal L^k}{\partial \mathbf z[j]} = \frac{\partial H \bigl({\mathbf p}^{k}, \mathbf y^k\bigr)}{\partial \mathbf z[j]} 
    & =\frac{-\partial \sum_{i=1}^{C_k}\mathbf y^k[i]\log \mathbf p^k[i]}{\partial \mathbf z[j]} \\
    & = - \frac{\partial \log \mathbf p^k[y^k]}{\partial \mathbf z[j]} \\
    & = -\frac{1}{\mathbf p^k[y^k]} \frac{\partial \mathbf p^k[y^k]}{\partial \mathbf z[j]}.
\end{aligned}
\end{equation}
Note that $y^k$ is the true label of the $k$-th hierarchy, and $\mathbf y^k$ is its corresponding one-hot label. 

According to Proposition 1, we have $\mathbf p^k=\mathbf T^{k\leftarrow 1}\mathbf p^1$. Thus,
\begin{equation}
\begin{aligned}
    \frac{\partial \mathbf p^k[y^k]}{\partial \mathbf z[j]} 
    & = \frac{\partial \mathbf (\mathbf T^{k\leftarrow 1}\mathbf p^1)[y^k]}{\partial \mathbf z[j]} \\
    & = \frac{\partial\sum_{n=1}^{C_1}t^{k\leftarrow1}_{y^kn}\mathbf p^1[n]}{\partial\mathbf z[j]} \\
    & = \sum_{n=1}^{C_1}t^{k\leftarrow1}_{y^kn} \frac{\partial \mathbf p^1[n]}{\partial\mathbf z[j]} \\
    & = t^{k\leftarrow1}_{y^kj}\frac{\partial\mathbf p^1[j]}{\partial \mathbf z[j]} + \sum_{n\neq j}t^{k\leftarrow1}_{y^kn} \frac{\partial \mathbf p^1[n]}{\partial\mathbf z[j]} \\
    & = t^{k\leftarrow1}_{y^kj} \mathbf p^1[j](1\!-\!\mathbf p^1[j]) \!+\! \sum_{n\neq j}t^{k\leftarrow1}_{y^kn}(-\mathbf p^1[n]\mathbf p^1[j]) \\
    & = t^{k\leftarrow1}_{y^kj} \mathbf p^1[j] + \sum_{n=1}^{C_1}t^{k\leftarrow1}_{y^kn}(-\mathbf p^1[n]\mathbf p^1[j]) \\
    & = t^{k\leftarrow1}_{y^kj} \mathbf p^1[j] - \mathbf p^1[j] \sum_{n=1}^{C_1}t^{k\leftarrow1}_{y^kn}(\mathbf p^1[n]) \\
    & = t^{k\leftarrow1}_{y^kj} \mathbf p^1[j] - \mathbf p^1[j] \mathbf p^k[y^k].
\end{aligned}
\end{equation}
Here, the last equation holds because $\mathbf p^k[y^k]=(\mathbf T^{k\leftarrow1}\mathbf p^1)[y^k] =\sum_{n=1}^{C_1}t^{k\leftarrow1}_{y^kn}(\mathbf p^1[n])$.

Combining Eq. (26) and Eq.(27), we obtain
\begin{equation}
\begin{aligned}
    \frac{\partial \mathcal L^k}{\partial \mathbf z[j]}
    & = -\frac{1}{\mathbf p^k[y^k]} \frac{\partial \mathbf p^k[y^k]}{\partial \mathbf z[j]} \\
    & = -\frac{1}{\mathbf p^k[y^k]} \bigl(t^{k\leftarrow1}_{y^kj} \mathbf p^1[j] - \mathbf p^1[j] \mathbf p^k[y^k]\bigr) \\
    & =\bigl(1-\frac{t^{k\leftarrow{1}}_{y^kj}}{\mathbf p^k[y^k]}\bigr)\mathbf p^1[j]
\end{aligned}
\end{equation}
This completes the proof.
\end{proof}

\subsection{Proof of Corollary 1}
\addtocounter{corollary}{-1}
\begin{corollary}
If the network parameters are updated according to Eq. (7) using SGD-based optimizers, then $\mathcal{G}^k_{j^*} < 0$ holds for $j^*=\arg\max_{j} t^{k\leftarrow{1}}_{y^k j}$, which is the subclass most visually similar to $y^k$ as determined by the Transition Networks.
\end{corollary}

\begin{proof}
Since $\mathbf p^k[y^k]=(\mathbf T^{k\leftarrow1}\mathbf p^1)[y^k] =\sum_{n=1}^{C_1}t^{k\leftarrow1}_{y^kn}(\mathbf p^1[n])$, we have
\begin{equation}
\label{eq:cor_1}
\begin{aligned}
    \mathcal{G}^k_{j^*} &= \bigl(1-\frac{t^{k\leftarrow{1}}_{y^k j^*}}{\mathbf p^k[y^k]}\bigr)\mathbf p^1[j^*] \\
    &= \bigl(1-\frac{t^{k\leftarrow{1}}_{y^k j^*}}{\sum_{n=1}^{C_1}t^{k\leftarrow1}_{y^kn}\mathbf p^1[n]}\bigr)\mathbf p^1[j^*] \\
    &= \bigl(1-\frac{t^{k\leftarrow{1}}_{y^k j^*}\sum_{n=1}^{C_1}\mathbf p^1[n]}{\sum_{n=1}^{C_1}t^{k\leftarrow1}_{y^kn}\mathbf p^1[n]}\bigr)\mathbf p^1[j^*] \\
    &= \bigl(1-\frac{\sum_{n=1}^{C_1}t^{k\leftarrow{1}}_{y^k j^*}\mathbf p^1[n]}{\sum_{n=1}^{C_1}t^{k\leftarrow1}_{y^kn}\mathbf p^1[n]}\bigr)\mathbf p^1[j^*],
\end{aligned}
\end{equation}
where the third equation holds due to $\sum_{n=1}^{C_1}\mathbf p^1[n]=1$.

According to $j^*=\arg\max_{j} t^{k\leftarrow{1}}_{y^k j}$, we obtain
\begin{equation}
    t^{k\leftarrow{1}}_{y^k j^*} \ge t^{k\leftarrow{1}}_{y^k n}, ~~~ n\in\{1,\cdots,C_1\}.
\end{equation}
Based on this inequality, it is easy to get that $\sum_{n=1}^{C_1}t^{k\leftarrow{1}}_{y^k j^*}\mathbf p^1[n] > \sum_{n=1}^{C_1}t^{k\leftarrow1}_{y^kn}\mathbf p^1[n]$. This implies that
\begin{equation}
    \bigl(1-\frac{\sum_{n=1}^{C_1}t^{k\leftarrow{1}}_{y^k j^*}\mathbf p^1[n]}{\sum_{n=1}^{C_1}t^{k\leftarrow1}_{y^kn}\mathbf p^1[n]}\bigr) < 0.
\end{equation}
Therefore, it can be concluded from \cref{eq:cor_1} that
\begin{equation}
    \mathcal{G}^k_{j^*} = \bigl(1-\frac{\sum_{n=1}^{C_1}t^{k\leftarrow{1}}_{y^k j^*}\mathbf p^1[n]}{\sum_{n=1}^{C_1}t^{k\leftarrow1}_{y^kn}\mathbf p^1[n]}\bigr)\mathbf p^1[j^*] < 0.
\end{equation}
This completes the proof.
\end{proof}

\subsection{Proof of Theorem 1}
In this section, we present a detailed proof of Theorem 1, which relies solely some mild and practical assumptions.
\addtocounter{theorem}{-1}
\begin{theorem}[]
\label{thr:distribution_alignment}
In missing label scenarios where the training sample $\mathbf x$ is partially annotated at the $k$-th hierarchy with ground-truth label $y^k$. If all entries of the feature vector input to the linear classification layer are non-zero \footnote{This assumption is practical, as we can introduce a small positive bias term (e.g, $\epsilon=10^{-8}$) or directly apply a \texttt{sigmoid} activation before the linear classification layer to ensure the input is greater than 0.}, and the network parameters are updated by $\mathcal L^k\triangleq H(\mathbf p^k, \mathbf y^k)$ using SGD-based optimizers, then at the minimum, the following equality holds:
\begin{equation}
    \tag{12}
    p(Y^1|\mathbf x;\omega) = p(Y^1|Y^k=y^k,\mathbf x;\omega,\theta),
\end{equation}
where $p(Y^1|\mathbf x;\omega)$ is formulated by the Classification Network with parameters $\omega$, and $p(Y^1|Y^k=y^k,\mathbf x;\omega,\theta)$ is jointly formulated by the Classification Network and the Transition Network with parameters $\omega$ and $\theta=\{\theta^1,\cdots,\theta^k\}$, respectively.
\end{theorem}

\begin{proof}
    By the extremum theorem in calculus, the minimums of $\mathcal L^k$ w.r.t the parameters $\omega$ of the Classification Network are achieved at $\frac{\partial \mathcal L^k}{\partial \omega}=0$. According to Proposition 2, we know that $\mathcal L^k$ is a function w.r.t the logits $\mathbf z$ of the finest-level hierarchy. Thus, we can unroll the gradient at $\mathbf z$ by applying the chain rule, which yields
    \begin{equation}
        \label{eq:thr_1}
        \underbrace{\frac{\partial L^k}{\partial \omega}}_{|w|\times 1} = \bigl(\underbrace{\frac{\partial \mathbf z}{\partial \omega}}_{C_1\times|\omega|}\bigr)^\top \underbrace{\frac{\partial\mathcal L^k}{\partial \mathbf z}}_{C_1\times 1},
    \end{equation}
    where we use $|\omega|$ to denote the dimensions of the parameters $\omega$. By using the conclusion of Proposition 2, we obtain
    \begin{equation}
        \frac{\partial \mathcal L^k}{\partial \mathbf z} = [\mathcal G^k_1,\cdots,\mathcal G^k_{C_1}]^\top,
    \end{equation}
    with
    \begin{equation}
        \label{eq:thr_2}
        \mathcal G^k_j=\frac{\partial \mathcal L^k}{\partial \mathbf z[j]}=(1 - \frac{t^{k\leftarrow{1}}_{y^k j}}{\mathbf{p}^k[y^k]}) \mathbf{p}^1[j],
    \end{equation}
    where $C_1$ denotes the category number of the finest-level hierarchy. Moreover, by utilizing the definition of transition probability in Eq. (3), $\mathcal G^k_j$ can be further expressed as:
    \begin{equation}
    \label{eq:thr_3}
    \begin{aligned}
        \mathcal G^k_j & =\bigl(1-\frac{t^{k\leftarrow{1}}_{y^k j}}{\mathbf p^k[y^k]}\bigr)\mathbf p^1[j] \\
        & = p(Y^1=j|\mathbf x)-\frac{p(Y^k=y^k|Y^1=j,\mathbf x)p(Y^1=j|\mathbf x)}{p(Y^k=y^k)} \\
        & = p(Y^1=j|\mathbf x)-p(Y^1=j|Y^k=y^k, \mathbf x),
    \end{aligned}
    \end{equation}
where the third equal sign holds due to Bayes' rule. 

Furthermore, we can substitute \cref{eq:thr_3} into \cref{eq:thr_2} to derive
\begin{equation}
\label{eq:thr_4}
\begin{aligned}
\frac{\partial \mathcal L^k}{\partial \mathbf z}  & \!=\!
\begin{bmatrix}
\mathcal G^k_1  \\
\mathcal G^k_2 \\
\vdots \\
\mathcal G^k_{C_1}
\end{bmatrix} \!=\!
\begin{bmatrix}
p(Y^1=1|\mathbf x)-p(Y^1=1|Y^k=y^k, \mathbf x) \\
p(Y^1=2|\mathbf x)-p(Y^1=2|Y^k=y^k, \mathbf x) \\
\vdots \\
p(Y^1=C_1|\mathbf x)-p(Y^1=C_1|Y^k=y^k, \mathbf x)
\end{bmatrix} \\
& \!=\!
\begin{bmatrix}
p(Y^1=1|\mathbf x) \\ p(Y^1=2|\mathbf x) \\ \vdots \\
p(Y^1=C_1|\mathbf x)
\end{bmatrix}
-
\begin{bmatrix}
p(Y^1=1|Y^k=y^k, \mathbf x) \\ p(Y^1=2|Y^k=y^k, \mathbf x) \\ \vdots \\ p(Y^1=C_1|Y^k=y^k, \mathbf x)
\end{bmatrix} \\
& = p(Y^1|\mathbf x) - p(Y^1|Y^k=y^k,\mathbf x).
\end{aligned}
\end{equation}
Note that, in the above derivation, we have omitted the parameters to simplify the notation. For example, $p(Y^1|\mathbf x;\omega)$ and $p(Y^1|Y^k=y^k,\mathbf x)$ correspond to $p(Y^1|\mathbf x)$ and $p(Y^1|Y^k=y^k,\mathbf x;\omega,\theta)$, respectively. From \cref{eq:thr_4}, it is evident that $\mathcal{G}^k_j$ inherently measures the distance between these two distribution.

By substituting \cref{eq:thr_4} to \cref{eq:thr_1}, we can obtain
\begin{equation}
    \label{eq:thr_5}
    \frac{\partial L^k}{\partial \omega} = \bigl(\frac{\partial \mathbf z}{\partial \omega}\bigr)^\top \bigl(p(Y^1|\mathbf x;\omega) - p(Y^1|Y^k=y^k,\mathbf x;\omega,\theta)\bigr),
\end{equation}

To complete the proof of the theorem, it is sufficient to show that $ p(Y_1 \mid x; \omega) - p(Y_1 \mid Y_k = y_k,\, x; \omega, \theta)=0$ holds when equation~\cref{eq:thr_5} is equal to zero, under the assumption that all entries of the feature vector input to the linear classification layer are non-zero. Note that $p(Y_1 \mid x; \omega) - p(Y_1 \mid Y_k = y_k,\, x; \omega, \theta) = 0$ is a sufficient condition for equation \cref{eq:thr_5} to be zero. Thus, to complete the proof, it suffices to verify that $ p(Y_1 \mid x; \omega) - p(Y_1 \mid Y_k = y_k,\, x; \omega, \theta)=0$ holds when the derivative of \( \mathcal{L}^k \) w.r.t a subset of the Classification Network’s parameters, such as the linear classification layer $W$, is zero under the same assumptions.

Without loss of generality, let us denote the features of $\mathbf x$ input to the linear classification layer as $\mathbf h\in\mathbb R^d$, such that the dimension of $W$ is $d \times C_1$. Consequently, the logits $\mathbf z$ of the finest-level hierarchy can be expressed as $\mathbf z=W^\top\mathbf h$. The derivative is then given by
\begin{equation}
\label{eq:thr_6}
\begin{aligned}
    \frac{\partial L^k}{\partial W} & = \bigl(\frac{\partial \mathbf z}{\partial W}\bigr)^\top \bigl(p(Y^1|\mathbf x;\omega) - p(Y^1|Y^k=y^k,\mathbf x;\omega,\theta)\bigr) \\
    & =\bigl(\frac{\partial W^\top\mathbf h}{\partial W}\bigr)^\top \bigl(p(Y^1|\mathbf x;\omega) - p(Y^1|Y^k=y^k,\mathbf x;\omega,\theta)\bigr) \\
    & =\mathbf h \bigl(p(Y^1|\mathbf x;\omega) - p(Y^1|Y^k=y^k,\mathbf x;\omega,\theta)\bigr)^\top.
\end{aligned}
\end{equation}

Now, we proceed to verify $ p(Y_1 \mid x; \omega) - p(Y_1 \mid Y_k = y_k,\, x; \omega, \theta)=0$ holds when the above equation achieves zero, under the assumption that all entries of the feature vector input to the linear classification layer are non-zero, i.e., $h_i\neq 0$ for all $i\in\{1,2,\cdots,d\}$.

Under the aforementioned assumption, we have that all entries of $\mathbf h$ are non-zero. Let $\frac{\partial L^k}{\partial W}=0$, it can then be readily observed that 
\begin{equation}
\begin{aligned}
    h_i \bigl(p(Y^1=j|\mathbf x;\omega) - p(Y^1=j|Y^k=y^k,\mathbf x;\omega,\theta)\bigr)=0, \\
    i\in \{1,2,\cdots,d\} \text{~~~and~~~} j\in\{1,2,\cdots,C_1\}.
\end{aligned}
\end{equation} 
must hold for each $i$ and $j$. Since $h_i\neq 0$ by assumption, we can conclude that $\bigl(p(Y^1=j|\mathbf x;\omega) - p(Y^1=j|Y^k=y^k,\mathbf x;\omega,\theta)\bigr)=0$ for all $j$. 

This further implies that $p(Y^1|\mathbf x;\omega) - p(Y^1|Y^k=y^k,\mathbf x;\omega,\theta)=0$ holds. Therefore, the proof is complete.
    
\end{proof}

\section{Why Learn Data-driven Transition Matrices?}

While the naive transition matrix in Eq. (10) can capture the pre-defined parent-child relationships, we would like to emphasize that additional predictions based on data-driven correlations, as employed in our work, are essential for effectively handling visual recognition tasks with hierarchical labels.

The main motivation for learning data-driven transition matrices is that the hierarchical relationships derived from a label hierarchy tree are not always visually grounded. In certain cases, such as for the Aircraft dataset, these relationships are defined more by semantics than by visual similarity. Specifically, the hierarchy in the Aircraft dataset is structured based on a maker-family-model organization (e.g., ``Boeing"–``Boeing 767"–``767-200"), but this semantic-based hierarchy does not always reflect visual similarity. For example:
\begin{itemize} 
    \item Boeing 737 (Boeing family) vs. Boeing 747 (Boeing family): Although both aircraft are produced by Boeing, the 737 is a narrow-body aircraft, whereas the 747 is a wide-body jumbo jet. The visual distinction between these models is quite significant, despite their shared maker. 
    
    \item Boeing 737 (Boeing family) vs. Airbus A320 (Airbus family): Although from different makers, both the 737 and A320 are narrow-body, twin-engine jets, making their visual similarities more evident than the comparison between the Boeing 737 and Boeing 747.
\end{itemize}
These examples illustrate that visual similarity does not always align with the hierarchical structure based on maker and family, underscoring the need for additional predictions to capture these visual correlations. 

To verify this, we compare the performance of the LTH\_naive model, which directly uses naive transition matrices, with our LHT model that employs learnable transition matrices, on the Aircraft dataset. As shown in \cref{fig:comparison_hist}, for the finest-level hierarchy, LTH\_naive achieves an ACC of 91.2, which is notably lower than the 92.6 achieved by our LHT model. To further demonstrate the importance of adaptively learning visual correlations, we constructed a new dataset, Perturbed-Aircraft, based on the Aircraft dataset. In this modified version, we retain the original three-level hierarchy but randomly perturb 40\% of the category indices at the $2$-nd hierarchy. This introduces significant disorder and misalignment in the category relationships, resulting in a label hierarchy that is less visually consistent than the original dataset Aircraft. Under this challenging scenario, the ACC of LHT\_naive at the finest-level hierarchy drops by about 4\%. In contrast, our LHT model surpasses LHT\_naive by 5.1\%, further demonstrating the effectiveness of learning data-driven transition matrices to capture visual category correlations.

The script and code for reproducing the experimental results on Perturbed-Aircraft have also been made publicly available at \href{https://github.com/renzhenwang/label-hierarchy-transition}{https://github.com/renzhenwang/label-hierarchy-transition}.

\begin{figure}[t]
    \centering
    \includegraphics[width=0.85\linewidth]{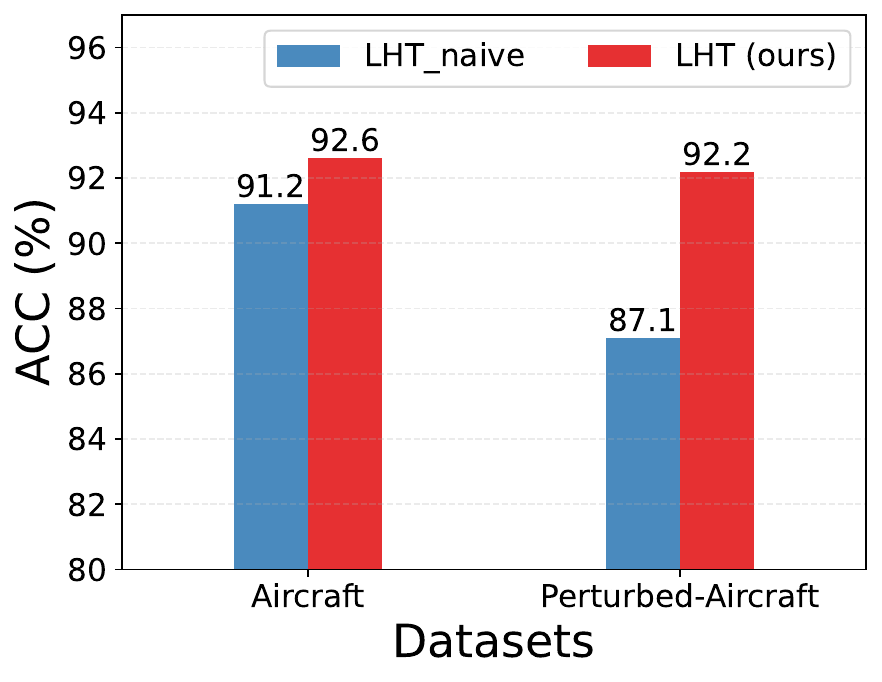}
    \caption{\revised{Performance comparison between LHT\_naive and LHT (ours) on the Aircraft and Perturbed-Aircraft datasets.}}
    \label{fig:comparison_hist}
    \vspace{4mm}
\end{figure}

\begin{figure*}[t]
\centering
\begin{minipage}[b]{0.48\linewidth}
    \centering
    \includegraphics[width=0.95\linewidth]{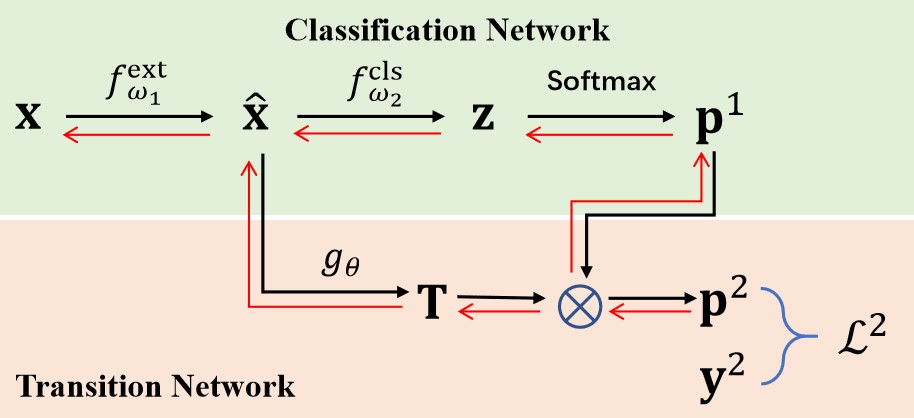}
    \caption*{(a) Fine-to-coarse Learning Paradigm}
\end{minipage}
\hfill
\begin{minipage}[b]{0.48\linewidth}
    \centering
    \includegraphics[width=0.95\linewidth]{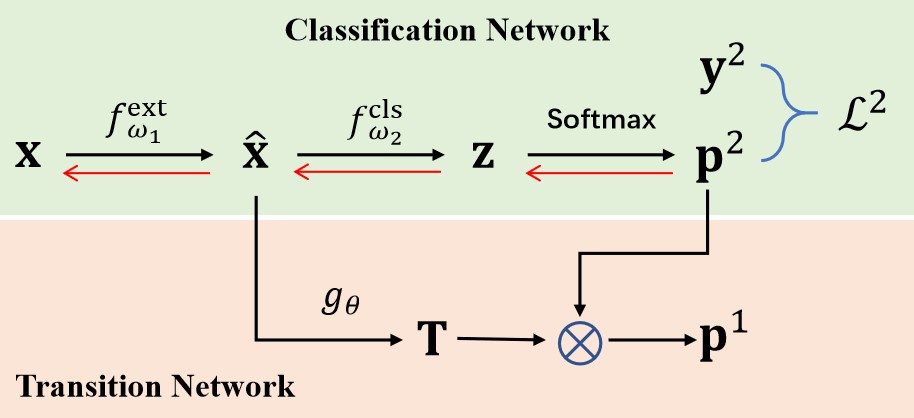}
    \caption*{(b) Coarse-to-fine Learning Paradigm}
\end{minipage}
\vspace{2mm}
\caption{\revised{Illustration of information flow during training for the sample $\mathbf x$ (labeled only at the 2-nd hierarchy $\mathbf y^2$). \textit{Notations}: $f_\omega=f^{\rm cls}_{\omega_1}\circ f^{\rm ext}_{\omega_2}$ denotes the Classification Network with parameters $\omega=\{\omega_1,\omega_2\}$, and $g_\theta$ represents the Transition Network with parameters $\theta$; Black lines represent data flow in the forward propagation process, while red lines indicate gradient flow in the backward propagation process.}}
\vspace{5mm}
\label{fig:flowchart}
\end{figure*}

\begin{table*}[t]
  \centering
  \caption{\revised{Comparison results on CUB-200-2011 dataset. $\mathcal R$ denotes that a fraction of $\mathcal R\%$ of the training samples lack labels at the finest-level (Species) hierarchy.}}
  \label{tab:c2f_vs_f2c}
  \resizebox{1\linewidth}{!}{
  \begin{tabular}{c |ccc| ccc| ccc| ccc| ccc}
    \toprule
    \multirow{2}{*}{Method} & \multicolumn{3}{c}{$\mathcal R=0$} & \multicolumn{3}{|c}{$\mathcal R=30$} & \multicolumn{3}{|c}{$\mathcal R=50$} & \multicolumn{3}{|c}{$\mathcal R=70$} & \multicolumn{3}{|c}{$\mathcal R=90$} \\
    \cmidrule(r){2-4}\cmidrule(r){5-7}\cmidrule(r){8-10}\cmidrule(r){11-13}\cmidrule(r){14-16}
    ~ & Order & Family & Sepcies & Order & Family & Sepcies & Order & Family & Sepcies & Order & Family & Sepcies & Order & Family & Sepcies \\
    \midrule
    LHT\_c2f & 98.9 & 95.9 & 86.6 & 98.8 & 95.6 & 84.7 & 98.8 & 95.3 & 82.4 & 98.9 & 94.9 & 78.3 & 98.7 & 94.8 & 65.1\\
    LHT\_f2c (ours) & 99.0 & 96.2 & 87.0 & 98.9 & 96.2 & 86.0 & 98.9 & 96.1 & 84.8 & 98.9 & 96.0 & 82.4 & 98.8 & 95.3 & 68.8 \\
    \bottomrule
  \end{tabular}
  }
\vspace{5mm}
\end{table*}

\section{Why not Learn the Transition Matrices form a Coarse-to-fine Manner?}

Technically, our LHT can learn the transition matrices in a fine-to-coarse manner, i.e., we first make predictions at the finest level, and the rest coarser-level labels can be determined according to the label hierarchy.We address the advantages of the proposed fine-to-coarse learning paradigm from the following three key aspects:

\begin{itemize}
    \item Motivationally, we employ an off-the-shelf classification network to predict labels of the finest-level hierarchy. Such a learning paradigm is advantageous in demonstrating that the Transition Network can be a plug-in to enhance conventional classification networks through hierarchical supervision, particularly in missing label scenarios where a large number of samples have labels only at partial coarser-level hierarchies.
    \item Theoretically, unlike the coarse-to-fine learning paradigm, the fine-to-coarse learning paradigm enables coarse-level hierarchies to provide supervision for the finest-level hierarchy via gradient back-propagation. This property is particularly important in missing label scenarios, as it allows for an effective utilization of partially labeled samples. To illustrate this, we take a two-level hierarchical classification as an example. As shown in Fig. \ref{fig:flowchart}(a), for a sample $\mathbf x$ labeled only at the second hierarchy $\mathbf y^2$, the fine-to-coarse learning paradigm can provide supervision to the finest-level hierarchy through $\mathbf y^2$. In contrast, this flexibility is not available in the coarse-to-fine paradigm, as shown in Fig. \ref{fig:flowchart}(b), where the gradient from the coarser-level label $\mathbf y^2$ cannot flow to the finest-level, thus failing to provide supervision for the finest-level hierarchy.
    
    \item Experimentally, we conducted experiments on the CUB-200-2011 dataset to compare the performance of the fine-to-coarse learning paradigm (LHT\_f2c) with that of the coarse-to-fine learning paradigm (LHT\_c2f). As shown in Table \ref{tab:c2f_vs_f2c}, the results demonstrate that LHT\_f2c consistently outperforms LHT\_c2f across various settings, with particularly significant improvements in missing label scenarios. This provides empirical evidence of the practical advantages of the fine-to-coarse paradigm in enhancing hierarchical classification.
\end{itemize}

\section{Delve into the hierarchical structure of the experimental benchmarks}
To clarify why CUB-200-2011, FGVC-Aircraft, and Stanford Cars are appropriate benchmarks for hierarchical classification, we provide qualitative and quantitative evidence that these datasets show clear inter-class similarity within hierarchies, making them well-suited for hierarchical learning.

\begin{figure*}[t] \vspace{0mm}
    \centering
    \includegraphics[width=0.85\linewidth]{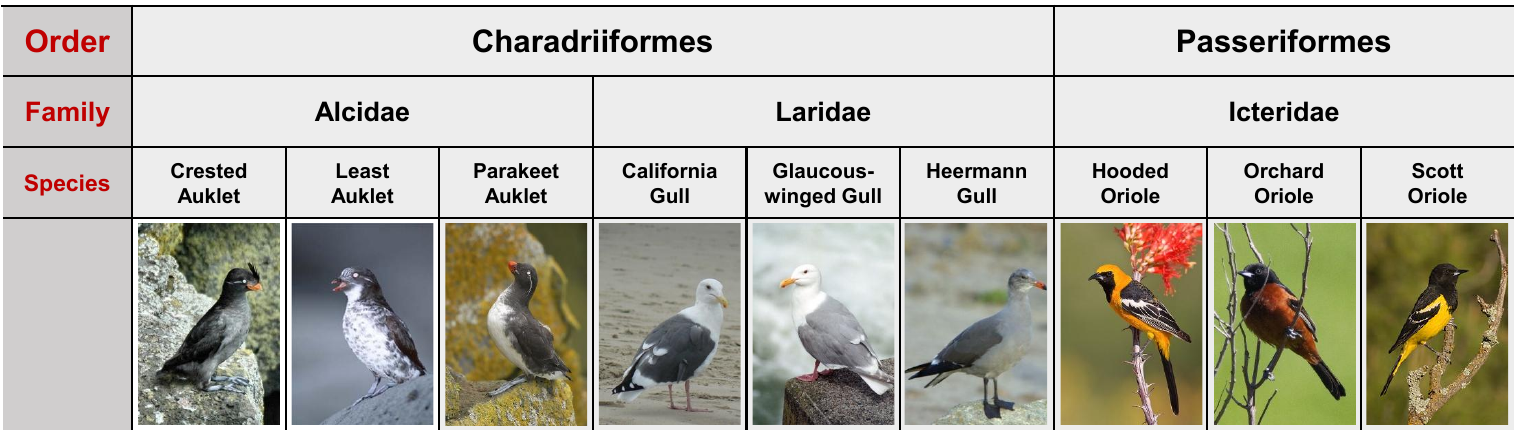}
    \vspace{0mm}
    \caption{Partial hierarchical structure in CUB-200-2011, where species are organized into three hierarchies (\textit{Species–Family–Order}). Birds exhibit strong visual similarity within the same hierarchy, but clear differences across higher-level ones.}
    \label{fig:cub}
    \vspace{0mm}
\end{figure*}

\begin{figure*}[t] \vspace{0mm}
    \centering
    \includegraphics[width=0.85\linewidth]{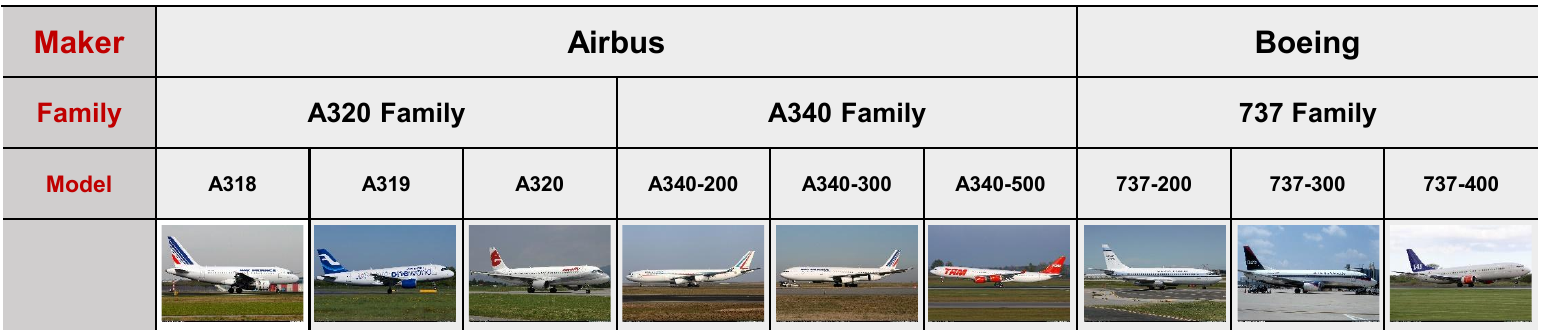}
    \vspace{0mm}
    \caption{Partial hierarchical structure in in FGVC-Aircraft, where models are organized into three hierarchies (\textit{Model–Family–Maker}). Aircraft exhibit strong visual similarity within the same hierarchy, but clear differences across higher-level ones.}
    \label{fig:air}
    \vspace{0mm}
\end{figure*}

\begin{figure}[H] \vspace{0mm}
    \centering
    \includegraphics[width=0.75\linewidth]{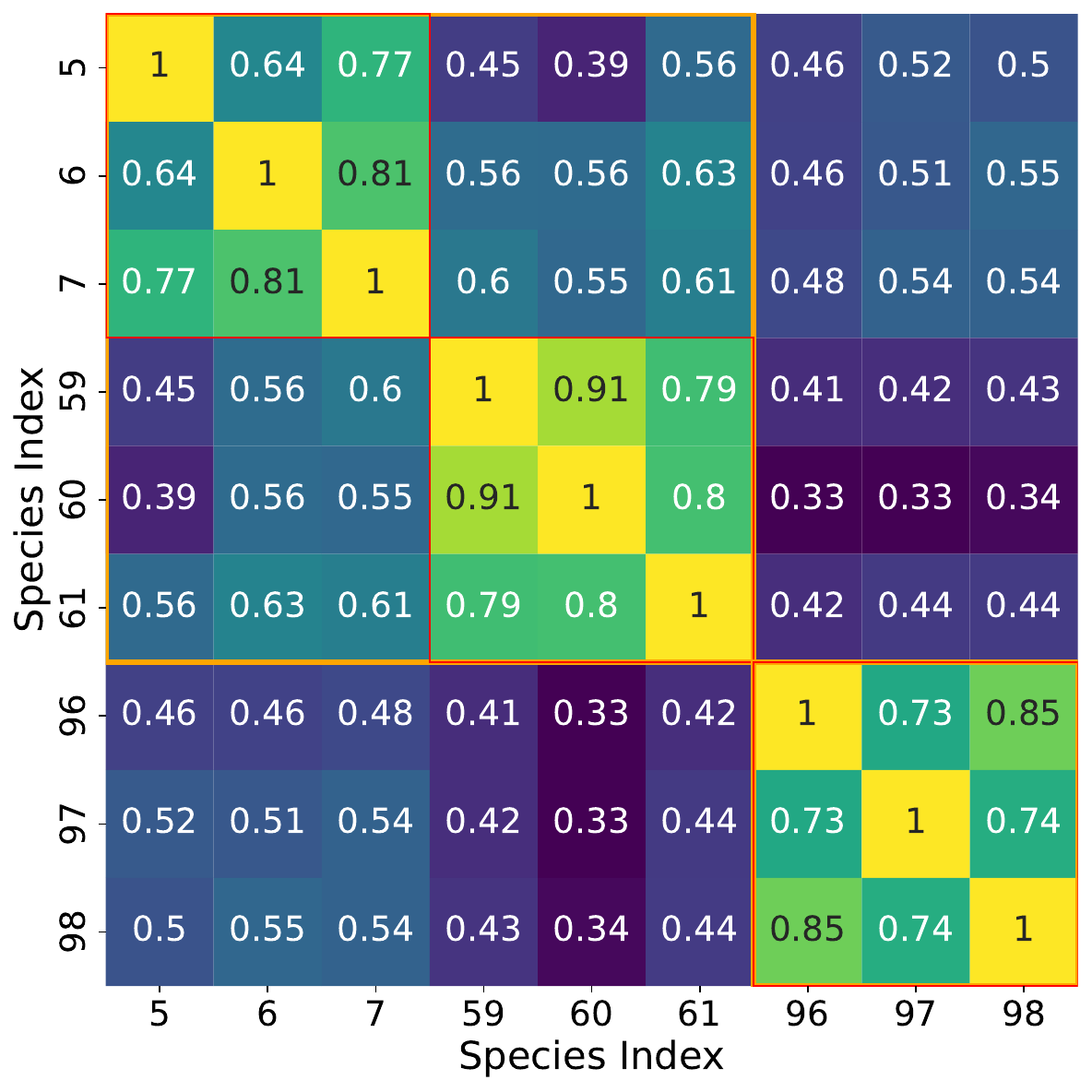}
    \vspace{0mm}
    \caption{Quantitative evidence of hierarchical structure in CUB-200-2011. Each entry in the similarity matrix represents the pairwise cosine similarity between two species. The matrix shows strong correlations within the same hierarchy, while differences become more apparent across higher-level ones.}
    \label{fig:similarity}
    \vspace{0mm}
\end{figure}

\begin{itemize}
    \item \textbf{Qualitative evidence supports the suitability of these datasets for hierarchical classification.} As shown in \cref{fig:cub} and \cref{fig:air}, the label hierarchies of these datasets align well with the underlying visual structures. In CUB, the taxonomy is grounded in biological knowledge, where families and orders naturally reflect visual traits such as plumage color and body shape. For example, the three auklet species within the Alcidae family are highly similar, whereas families such as Alcidae and Passerellidae exhibit obvious distinctions. In FGVC-Aircraft, the hierarchy is artificially defined by a Maker–Family–Model structure, yet it still encodes meaningful visual correlations. For example, models such as Airbus A318, A319, and A320 are visually similar, whereas cross-family differences (e.g., Airbus vs. Boeing) are more obvious. Although Stanford Cars provides only a two-level hierarchy, it still follows the same principle as FGVC-Aircraft, where the models within a maker are highly similar. Overall, these datasets demonstrate that their label trees effectively encode hierarchical visual correlations, making them well-suited for the evaluation of hierarchical classification methods.
    
    \item \textbf{Quantitative evidence supports the suitability of these datasets for hierarchical classification.} To substantiate the above claims, we further conduct a quantitative analysis on CUB using its rich attribute annotations. Each species in CUB is associated with a 312-dimensional attribute vector that encodes detailed properties of color, shape, pattern, and specific body parts (e.g., wing color, bill shape, underpart pattern, et al.). We compute the pairwise cosine similarity between these vectors by $s_{ij} = \frac{\langle \mathbf a_i, \mathbf a_j \rangle}{\|\mathbf a_i\|\|\mathbf a_j\|}$, where $\mathbf a_i$ denotes the attribute vector of the species $c_i$. The resulting similarity matrix is shown in \cref{fig:similarity}, and it should be noted that the axes correspond to the species indices, consistent with those in \cref{fig:cub}. The matrix exhibits clear block-diagonal structures aligned with taxonomic structures. In particular, species within the same family (e.g., Alcidae or Icteridae) exhibit substantially higher similarities than those across different families. These results indicate that the given label tree closely corresponds to actual semantic and visual similarities between species, providing a solid foundation for evaluating hierarchical classification.
\end{itemize}

\section{Visualization of the value of the learned weight $\gamma$.}

\begin{figure}{r}
    \centering
    \includegraphics[width=0.35\textwidth]{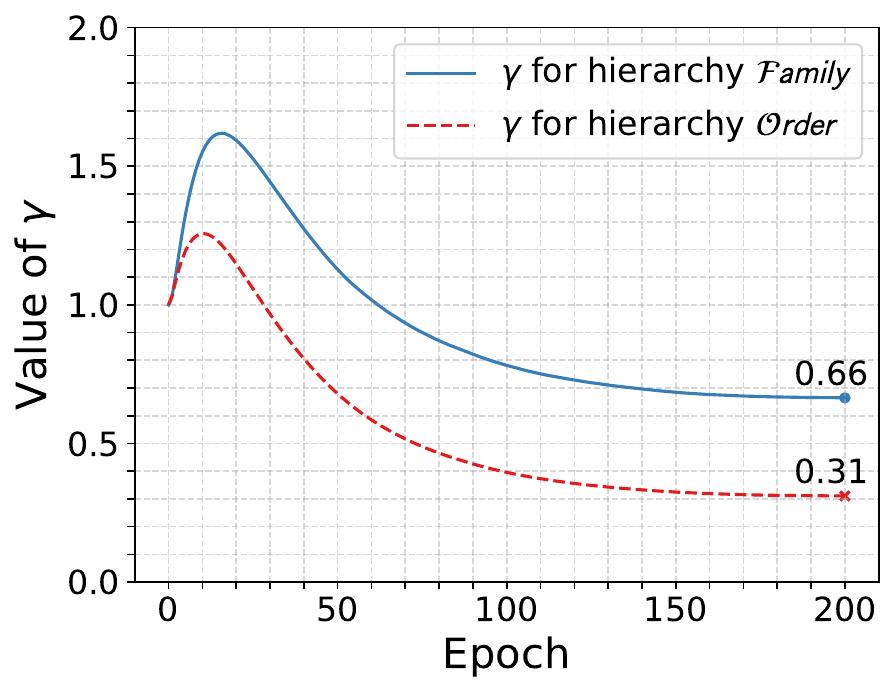}
    \caption{\revised{Learning curves of $\gamma$ during the training stage on CUB-200-2011 with $\mathcal R=0\%$ (corresponding to the setup of Fig. 6 in the original manuscript)}.}
    \label{fig:gamma}
\end{figure}

To provide a clearer understanding of the impact of the learned transition matrices, we here visualize the evolution of the weight $\gamma$ throughout the training stage on the CUB-200-2011 dataset with $\mathcal{R}=0\%$ (corresponding to the setup of Fig. 6 in the original manuscript) in \cref{fig:gamma}. It can be observed that: (1) By the end of training, $\gamma$ deviates significantly from its initial value of 1.0 and exhibits distinct values across different hierarchies, indicating that the learned components differ substantially from the naive transition matrices; (2) Throughout the training process, $\gamma$ shows dynamic changes, which supports our initial motivation that $\gamma$ helps maintain the values of $\gamma\mathbf S$ and the transition network outputs on the same scale. Such an operation is necessary because the elements of $\mathbf S$ are constrained within the range of 0 to 1, while the outputs of the transition network are unconstrained. Scaling $\mathbf S$ by a learnable $\gamma$ helps ensure numerical stability during network training.

\begin{table}[h]
  \renewcommand\arraystretch{1.1}
  \centering
  \caption{Comparison with FGoN \cite{chang2021your} across different datasets ($\mathcal R=100\%$) under the low-resolution input settings. $^*$ indicates the results reported in the original paper.}
  \label{tab:rescale}
  \setlength{\tabcolsep}{2.2mm}{
  \begin{tabular}{c c cc cc}
    \toprule
    \multirow{2}{*}{Dataset} & \multirow{2}{*}{Hierarchy} & \multicolumn{2}{c}{FGoN$^{*}$} & \multicolumn{2}{c}{Ours} \\
    \cmidrule(r){3-4} \cmidrule(r){5-6}
    ~ & ~ & ACC & mACC & ACC & mACC \\
    \midrule
    \multirow{3}{*}{Cub-200-2011} & Order & 96.4 & \multirow{3}{*}{88.2} & 97.7 & \multirow{3}{*}{90.0}  \\
    ~ & Family & 90.4 &  & 92.3 &  \\
    ~ & Species & 78.0 &  & 79.9 & \\
    \hline
    \multirow{3}{*}{Aircraft} & Maker & 93.0 & \multirow{3}{*}{90.7} & 95.6 & \multirow{3}{*}{92.8}  \\
    ~ & Family & 90.7 &  & 93.4 &  \\
    ~ & Model & 88.4 &  & 89.4 & \\
    \hline
    \multirow{2}{*}{Stanford Cars} & Type & 95.6 & \multirow{2}{*}{92.7} & 96.1 & \multirow{2}{*}{93.3}  \\
    ~ & Maker & 89.7 &  & 90.4 &  \\
    \bottomrule
  \end{tabular}
  }
\end{table}

\section{How does LHT perform when meeting low-resolution images?} 
As investigated in \cite{chang2021your}, low-resolution images may potentially degrade the performance of a hierarchical classification system, and the proposed FGoN \cite{chang2021your} shows evident superiority to take advantage of hierarchical label information, thus mitigating the harm brought about by the image quality. To validate the effectiveness of the proposed method in low-resolution images settings, we directly compare with FGoN following the same training protocols. Concretely, we reduce the image resolution to $224\times224$, and keep other settings unchanged for fair comparison. As shown in \cref{tab:rescale}, our proposed LHT model consistently surpasses FGoN, and achieves absolute mACC gains of 1.8\%, 2.1\% and 0.6\% on Cub-200-2011, Aircraft and Stanford Cars datasets, respectively. This verifies that our method has fine potential to infer category labels from class hierarchies, so as to alleviate the practical limitations arising from image quality.


%
\bibliographystyle{IEEEtran}
\bibliography{lht}

\ifCLASSOPTIONcaptionsoff
  \newpage
\fi

\end{document}